\pdfoutput=1
\documentclass[11pt]{article}

\usepackage[final]{acl}
\usepackage{fdsymbol}

\usepackage{times}
\usepackage{latexsym}
\usepackage{makecell}
\usepackage{tablefootnote}
\usepackage[toc,page]{appendix}
\usepackage{pifont}% http://ctan.org/pkg/pifont
\newcommand{\cmark}{\ding{51}}%
\newcommand{\xmark}{\ding{55}}%
\usepackage{algorithm2e}
\usepackage{algpseudocode}
\usepackage{enumitem}
\usepackage{fancyvrb}
\usepackage{cuted, tcolorbox}
\usepackage[T1]{fontenc}
\usepackage[utf8]{inputenc}

\usepackage{microtype}

\usepackage{hyperref}
\usepackage{booktabs}
\usepackage{graphicx}
\usepackage{makecell}
\graphicspath{{./imgs/}}
\usepackage{footmisc}
\usepackage{alltt}
\usepackage{floatrow}
\usepackage{arydshln}
\usepackage{microtype}

\usepackage{caption}
\usepackage{subcaption}
\usepackage{array, makecell} %
\usepackage{amsmath}
\usepackage{pifont}% http://ctan.org/pkg/pifont
\usepackage{tabularx,colortbl}

\usepackage{tikz}
\usepackage{collcell}

\usepackage{etoolbox}

\newcolumntype{?}{!{\vrule width 1.5pt}}

\newtoggle{inTableHeader}% Track if still in header of table
\toggletrue{inTableHeader}% Set initial value
\newcommand*{\StartTableHeader}{\global\toggletrue{inTableHeader}}%
\let\OldTabular\tabular%
\let\OldEndTabular\endtabular%
\renewenvironment{tabular}{\StartTableHeader\OldTabular}{\OldEndTabular\StartTableHeader}%

\newcommand*{\MinNumber}{-1.0}%
\newcommand*{\MidNumber}{0.0} %
\newcommand*{\MaxNumber}{1.0}%

\newcommand{\ApplyGradient}[1]{%
  \iftoggle{inTableHeader}{#1}{
    \ifdim #1 pt > \MidNumber pt
        \pgfmathsetmacro{\PercentColor}{max(min(100.0*(#1 - \MidNumber)/(\MaxNumber-\MidNumber),100.0),0.00)} %
        \hspace{-0.33em}\colorbox{yellow!\PercentColor!blue}{#1}
    \else
        \pgfmathsetmacro{\PercentColor}{max(min(100.0*(\MidNumber - #1)/(\MidNumber-\MinNumber),100.0),0.00)} %
        \hspace{-0.33em}\colorbox{blue!\PercentColor!blue}{#1}
    \fi
  }}
\newcolumntype{R}{>{\collectcell\ApplyGradient}c<{\endcollectcell}}

\usepackage{amsmath}
\usepackage{amsfonts,bm}
\usepackage{xspace}

\newcommand{\Ni}{({\em i})~}
\newcommand{\Nii}{({\em ii})~}
\newcommand{\Niii}{({\em iii})~}

\definecolor{mypink3}{cmyk}{0, 0.7808, 0.4429, 0.1412}

\makeatletter   
\newcommand{\sveryshortarrow}[1][3pt]{\mathrel{%
    \vcenter{\hbox{\rule[-.5\fontdimen8\scriptfont3]
               {\scriptratio\dimexpr#1\relax}{\fontdimen8\scriptfont3}}}%
   \mkern-4mu\hbox{\let\f@size\sf@size\usefont{U}{lasy}{m}{n}\symbol{41}}}}
\makeatother

\def\eqref#1{equation~\ref{#1}}
\def\1{\bm{1}}

\def\m1{{\bm{1}}}

\DeclareMathAlphabet{\mathsfit}{\encodingdefault}{\sfdefault}{m}{sl}
\SetMathAlphabet{\mathsfit}{bold}{\encodingdefault}{\sfdefault}{bx}{n}

\usepackage[nameinlink]{cleveref}
\crefformat{section}{\S#2#1#3} % see manual of cleveref, section 8.2.1
\crefname{algorithm}{Alg.}{Algs.}
\crefformat{subsection}{\S#2#1#3}
\Crefname{equation}{Eq.}{Eqs.}
\Crefname{figure}{Fig.}{Figs.}

\usepackage[colorinlistoftodos,prependcaption,textsize=tiny]{todonotes}

\usepackage{soul}

\usepackage{float}
\definecolor{azure}{rgb}{0.0, 0.5, 1.0}
\definecolor{darkbrown}{rgb}{0.4, 0.26, 0.13}
\definecolor{moonstoneblue}{rgb}{0.45, 0.66, 0.76}
\usepackage{multirow}% http://ctan.org/pkg/multirow
\usepackage{multicol}
\usepackage{hhline}% http://ctan.org/pkg/hhline

\newcommand{\model}{{\textsc{\textbf{DataNarrative}}}}

\title{{\model}:  Automated Data-Driven Storytelling  with \\Visualizations and Texts}

\author{
Mohammed Saidul Islam$^{\clubsuit}$, \ Md Tahmid Rahman Laskar$^{\clubsuit\heartsuit}$ \\ \bf \ Md Rizwan Parvez$^{\varheartsuit}$, \ Enamul Hoque$^{\clubsuit}$, \ Shafiq Joty$^{\vardiamondsuit\spadesuit}$\\
$^\clubsuit$York University, Canada, $^\spadesuit$Nanyang Technological University, Singapore, \\
$^\heartsuit$Dialpad Inc., Canada,
$^\varheartsuit$Qatar Computing Research Institute (QCRI),  $^\vardiamondsuit$Salesforce AI \\
\{saidulis, enamulh, tahmid20\}@yorku.ca, mparvez@hbku.edu.qa, sjoty@salesforce.com 
}

\begin{document}
\maketitle

\begin{abstract} 
Data-driven storytelling is a powerful method for conveying insights by combining narrative techniques with visualizations and text. These stories integrate visual aids, such as highlighted bars and lines in charts, along with textual annotations explaining insights. However, creating such stories requires a deep understanding of the data and meticulous narrative planning, often necessitating human intervention, which can be time-consuming and mentally taxing. While Large Language Models (LLMs) excel in various NLP tasks, their ability to generate coherent and comprehensive data stories remains underexplored. In this work, we introduce a novel task for data story generation and a benchmark containing 1,449 stories from diverse sources. To address the challenges of crafting coherent data stories, we propose a multi-agent framework employing two LLM agents designed to replicate the human storytelling process:  
one for understanding and describing the data (Reflection), generating the outline, and narration and another for verification at each intermediary step.
While our agentic framework generally outperforms non-agentic counterparts in both model-based and human evaluations, the results also reveal unique challenges in data story generation.

\end{abstract}

\section{Introduction}

\begin{figure}[t]
     \centering
\includegraphics[width=.94\columnwidth]{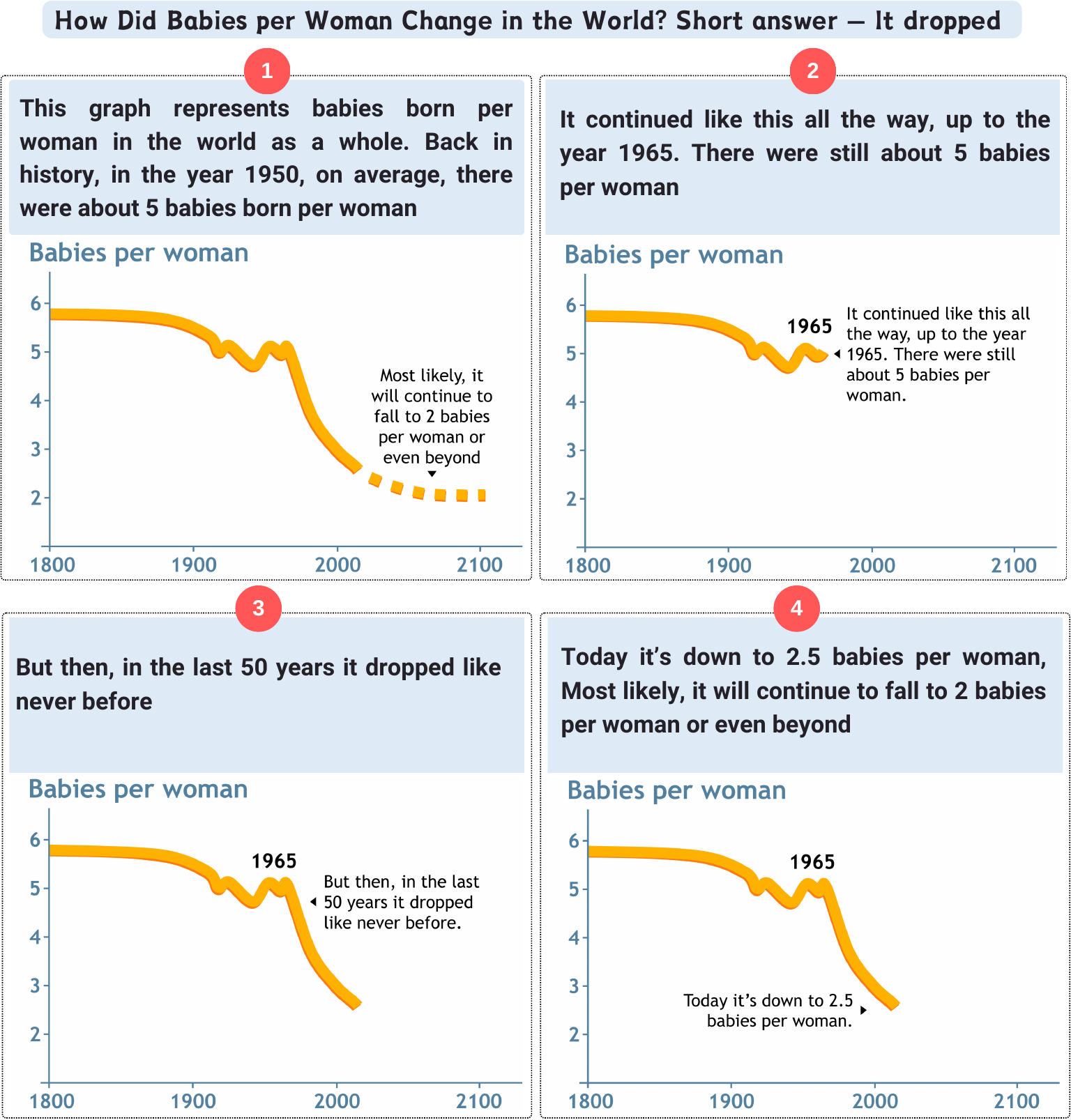}
         \caption{
        An example data story in our corpus extracted from GapMinder \cite{gapminder}             
          }
\label{fig:gap_data_story}
\end{figure}

Visual data stories have emerged as a powerful medium for communicating data, effectively combining the strengths of visualizations and text to convey contextual information and causal relationships~\cite{hullman2011visualization}. Ranging from data scientists to business analysts to journalists, people frequently write data-driven reports that integrate charts and text to present information to readers in a clear, coherent
and visually engaging manner~\cite{otten2015infographics}.
The essence of a visual data story involves identifying compelling insights within data (``story pieces''), presenting them through visualizations and texts, and arranging these representations into a coherent narrative that communicates an overarching message \cite{lee2015morethantelling}. Well-crafted visual stories have the potential to significantly enhance data understanding, even for those without specialized technical backgrounds. By combining narrative with data visualization, authors can illustrate trends, highlight correlations, and uncover hidden insights that might be lost in dense tables or reports.  For example,  \Cref{fig:gap_data_story} shows a GapMinder data story \cite{gapminder} in which renowned storyteller \href{https://en.wikipedia.org/wiki/Hans_Rosling}{Hans Rosling} explained how birth rates in the world have changed over time using text and charts.  

Data storytelling is widely used across various companies, including Microsoft and Tableau to effectively communicate insights and drive decision-making. In business intelligence, it can help present sales trends and performance metrics, while in healthcare, it can help illustrate patient outcomes and track disease outbreaks. Marketers can leverage it to optimize strategies through customer behavior visualization, and financial analysts use it for investment performance and risk assessments. In education, it can help track students' performance highlighting areas where they excel and where they might need additional support, and in public policy, it can communicate the impact of policies on social issues, for instance, a data story could show how a new housing policy affected homelessness rates, providing evidence-based insights to policymakers and the public alike.

Despite the popularity of data-driven stories, crafting them remains challenging and time-consuming, requiring skills in data analysis, visualization, graphic design, and storytelling. 
To facilitate data-driven storytelling, extensive research has introduced new concepts, theories, and tools. For instance, 
\citet{segel2010narrative} explored different design spaces from a narrative structure point of view, while others \cite{hullman2013adeeper, lan2022kinecharts, mckenna2017visualnarrative, shi2021communicating, shi2021understanding} focused on visual representations for crafting visual stories, tailoring their approaches based on specific tasks and communication objectives. While insightful and coherent, manually created data stories require significant human effort and time. In response, efforts have been made to develop automated methods for generating data stories \cite{shi2019taskoriented, shi2021calliope, wang2020datashot}, but these often produce simple facts lacking in quality and engaging narratives.

The rise of LLMs has prompted researchers to explore their effectiveness in tasks like chart summarization \cite{kantharaj-etal-2022-chart,rahman2023chartsumm}, chart question answering \cite{masry-etal-2022-chartqa, kantharaj-etal-2022-opencqa} and natural language story generation \cite{zhou2023recurrentgpt, xie-riedl-2024-creating}. However, the ability of LLMs to generate stories from data tables and to understand their effectiveness remains largely unexplored partly  
because of the lack of a benchmark dataset.

To address the research gap, we first develop a new task and the corresponding benchmark consisting of 1,449 data stories collected from real-world sources.. 
Motivated by the impressive performance of LLM-based agents in various planning tasks~\cite{ge2023openagi, yang2023autogpt, wang2023voyager, modarressi2023retllm, chen2024autoagents, wu2023autogen}, we then propose an agentic framework which takes data tables as inputs and employs 
two LLM agents -- a Generator or Actor 
and an Evaluator 
or Critic -- to mimic the human process of data story generation through writing and revising based on Critic's feedback 
(Figure~\ref{fig:agentic_framework}). 
The process includes a planning step (reflection and outline generation) and a story generation step (narration), with each step verified and revised by the critic LLM, creating a feedback loop to ensure coherence and factual consistency. Experimental results show that our agentic framework 
outperforms
non-agentic LLM counterparts in terms of generating more insightful and coherent stories with better resemblance to human-written narratives. 

Our main contributions include: 
\Ni a new automatic data story generation task and a corresponding benchmark dataset, 
\Nii a multi-step LLM-agent framework for  Data Story Generation. \Niii extensive automatic and human evaluations that demonstrate the state-of-the-art performance of \model. 
We make our code and data story corpus publicly available at \href{https://github.com/saidul-islam98/DataNarrative}{here.}

\section{Related Work}
\subsection{Story Generation Tasks}
Automated story generation is an open-ended task focusing on generating a sequence of events based on specific criteria~\cite{li2013story}. Generated stories  can be textual~\cite{kumar2006algorithms}, visual~\cite{li2019storygan, cohn2020visual}, or multimodal~\cite{bensaid2021fairytailor}. Visual stories, often found in comics and storyboards, present image sequences centered around main characters~\cite{cohn2020visual}. Early visual story generation models 
primarily utilized either global image features \cite{yu-etal-2017-hierarchically, wang-etal-2018-metrics, Huang_Gan_Celikyilmaz_Wu_Wang_He_2019} or local features, which focus on specific parts of an image, such as objects \cite{Wang_Wei_Li_Zhang_Huang_2020, hong-etal-2020-diverse, braude2022ordered}, to create visually grounded stories.

Data-driven stories differ from visual stories as they produce multimodal outputs in which charts communicate patterns, trends, and outliers in data while text explains such visualizations~\cite{riche2018data, kwon2014visjockey, segel2010narrative, hullman2013contextifier}. Early work focused on extracting and ranking key insights from data tables using statistical measures~\cite{rui2019quickinsights, bo2017extracting}. Tools like DataShot~\cite{wang2020datashot} and Calliope~\cite{shi2021calliope} present data facts with visualizations and captions, while Erato~\cite{sun2023erato} and Socrates~\cite{wu2024socrates} incorporate user input to guide the story generation process. In addition, there has been a recent survey \cite{he2024leveraging} that explores the utilization of large models in narrative visualization. However, the methods used in existing works often use simple rule-based approaches that may miss critical insights and lack effective narrative structure.

\subsection{LLMs for Story Generation}

Recent LLMs such as Gemini~\cite{geminiteam2023gemini}, ChatGPT~\cite{Chatgpt}, and GPT-4~\cite{GPT4} excel at generating fluent stories by repeatedly providing contextual information from both the plan and the current state of the story to an LLM prompt~\cite{yang-etal-2022-re3, wang-etal-2023-improving-pacing}. Several studies confirm the effectiveness of LLMs in generating short~\cite{eldan2023tinystories}, coherent and fluent stories~\cite{peng-etal-2022-inferring}. However, data story generation using LLMs is rare; one exception is DataTales~\cite{sultanum2023datatales}, which uses LLMs for narrative generation from chart images but is limited to only producing textual narratives without charts.

Recent studies also explore LLM agents in decision-making~\cite{yang2023autogpt}, task planning in video games~\cite{wang2023voyager}, memory function configuration~\cite{modarressi2023retllm}, multi-agent conversations~\cite{wu2023autogen}, and code generation~\cite{ridnik2024code, islam2024mapcoder}. Despite the suitability of this approach for open-ended tasks requiring planning, 
LLM agents for data story generation remain unexplored.

\subsection{Chart-related Downstream Tasks}
Several downstream tasks associated with charts have been proposed recently. \citet{masry-etal-2022-chartqa, methani2020plotqa} focus on answering factual questions about charts that require arithmetic and visual reasoning, while \citet{kantharaj-etal-2022-opencqa} address open-ended question-answering that generates explanatory texts. Chart summarization task involves generating informative summaries from a chart \cite{kantharaj-etal-2022-chart, tang2023vistext,rahman2023chartsumm}, while Chart-to-Table \cite{choi2019visualizing, masry2023unichart, masry2024chartinstruct} extracts the underlying data tables from a chart image. Others focus on verifying claims about charts~\cite{akhtar-etal-2023-reading, akhtar2024chartcheck}. Unlike the above tasks which produce only text, data-driven stories are multimodal as they combine visualizations with texts and there are no existing benchmarks for this task.

\section{Benchmark Construction}

Given the lack of a benchmark for automated data storytelling, we started by exhaustively searching across diverse online sources such as news sites, visualization repositories, and data blog sites.  At the end, we chose three suitable sources that contain data stories covering a series of visualizations and texts as we described below.

\subsection{Data Collection}
\noindent{$\bullet$} \, \textbf{Pew} \, \, 
Pew Research \cite{pewresearch} publishes data reports related to social issues, public opinion, and demographic trends. Often, such reports include charts and accompanying texts to communicate a coherent data story. To assemble the Pew corpus, we crawled articles from the Pew Research website until March 14, 2024, resulting in 4,532 articles across 18 topics and 22,760 figures (i.e., charts and other images). For each article, we extracted the title, paragraphs, and chart images and their metadata (e.g., captions and alt-texts).

\begin{table}[t]
\centering
\caption{Distribution of stories, charts, and tables across the train and test split 
of three datasets}
\label{tab:train-test}
\resizebox{\columnwidth}{!}{%
\begin{tabular}{lcc|cc|cc}
\multirow{2}{*}{} & \multicolumn{2}{c}{\textbf{Pew}} & \multicolumn{2}{c}{\textbf{Tableau}} & \multicolumn{2}{c}{\textbf{GapMinder}} \\ \midrule
       \# of Samples                & Train       & Test      & Train         & Test        & Train          & Test         \\ \midrule
\# of Stories             & 1,068        & 321       & 42            & 13          & -              & 5            \\
\# of Tables            & 4,729        & 1,590      & 340           & 64          & -              & 42           \\
\# of Charts            & 4,729        & 1,590      & 297           & 64          & -              & 42       \\\Xhline{1pt}   
\end{tabular}%
}
\end{table}
\noindent{$\bullet$} \, \textbf{Tableau} \, \, 
Tableau Public Story \cite{tableaupublic} allows users to create interactive stories through data visualizations on various topics and make these stories publicly accessible. Collecting data from Tableau with web crawlers proved difficult due to the complicated nature of the story representation, leading us to manually curate stories from the website. Specifically, we looked for stories that presented a paginated view, each page containing text and an associated chart. We searched by terms like `story', `data story', and `narrative-visualization' on the Tableau public, which led us to find over 1,200 dashboards with potential data stories. From these, we filtered out dashboards that did not have paginated views with a series of pages containing both text and charts. This filtering process led us to select 100 candidate stories for our corpus.  For each story page, we downloaded the chart image, data table, title, and text. 

\noindent{$\bullet$} \, \textbf{GapMinder} \, \, 
GapMinder \cite{gapminder} offers interactive data visualization tools and educational resources on global trends in health, wealth, and development indicators. Similar to Tableau stories, GapMinder stories were challenging to crawl due to the tool's interactive nature. Additionally, only a small subset of data articles featured both a paginated view and a combination of text and charts, resulting in 11 data stories. For each page in these stories, we downloaded the chart image and other associated data. 

\subsection{Data Processing \& Annotation}
Data processing and annotations follow three steps: \textit{(i)} story filtering, \textit{(ii)} chart data extraction, \textit{(iii)} chart-text pairs identification. 

\noindent{$\bullet$} \, \textbf{Story Filtering} \, \, To ensure the quality of our corpus, we applied the following exclusion criteria (\textbf{EC}) for filtering data stories from the initial collection: \Ni stories with texts shorter than 500 tokens for Pew and 140 tokens for Tableau and GapMinder samples, \Nii  Stories with fewer than 3 or more than 10 charts.
\begin{table}[t]
\centering
\caption{Chart type distribution}
\label{tab:chart-type}
\resizebox{\columnwidth}{!}{%
\begin{tabular}{lcc|cc|cc}
\multirow{2}{*}{} & \multicolumn{2}{c}{\textbf{Pew}} & \multicolumn{2}{c}{\textbf{Tableau}} & \multicolumn{2}{c}{\textbf{GapMinder}} \\\midrule
    Type & Train & Test & Train & Test & Train & Test \\ \midrule
Bar     & 3949  & 1159 & 155   & 46   & -     & -    \\
Line    & 433   & 360  & 69    & 8    & -     & 31   \\
Pie     & 191   & 53   & 9     & 2    & -     & -    \\
Scatter & 42    & 10   & 36    & 6    & -     & -   \\
Bubble & -    & -   & 16    & 1    & -     & 11   \\
Other   & 114   & 8    & 12    & 1    & -     & -    \\ \midrule
Total   & 4729  & 1590 & 297   & 64   & -     & 42  \\\Xhline{1pt} 
\end{tabular}%
}
\end{table}
By applying these  criteria, we carefully selected 
the stories from Pew, Tableau, and GapMinder, 
resulting in a total of 1,449 stories. Also, some Tableau stories included complex and unconventional visualizations, such as infographics and treemaps, so we filtered these stories to retain the ones with common visualizations.

\noindent{$\bullet$} \, \textbf{Chart data extraction} \, \, Chart data tables are essential for the story-generation process as we use them as inputs to the proposed framework. Also, to identify the text associated with each chart, we first need to extract the underlying data table of the chart image. We managed to download some gold data tables either from the story page (for Tableau) or from external sources (\citet{owid} for Gapminder). However, for Pew, we needed to automatically extract data from chart images as the original data tables were not available.  
Specifically, we utilized the multi-modal large language model Gemini-1.0-pro-vision \cite{geminiteam2023gemini} to extract data from chart images, which has been found to be effective for this task~\cite{islam2024large}. On 100 chart images from the ChartQA \cite{masry-etal-2022-chartqa} corpus, where gold tables were already available, we manually evaluated and found that the model correctly generated the tables in 77\% of the cases (more details in  \Cref{app:chart_data_ext}). 

\noindent{$\bullet$} \, \textbf{Identification of  chart-text pairs} \, \,
Since data stories usually come with descriptive texts for charts, it was essential to identify the texts related to each chart. Given the relatively small sizes of the Tableau and GapMinder corpus, we manually extracted the paragraphs associated with each chart image. For Pew, the chart-text pairs were already identified in the Chart-to-Text corpus \cite{kantharaj-etal-2022-chart} for 321 articles. However, for the remaining 1068 articles, we did not have the chart-text pairs. Due to the large sample size, collecting chart-text manually would be labor-intensive and time-consuming. Therefore, we utilized the state-of-the-art GPT-4-turbo model \cite{gpt4turbo} to collect relevant paragraphs corresponding to each of the charts in the training set.  On a small subset of human-annotated Chart-to-Text corpus, the model accurately linked paragraphs to data tables 70\% of the time (more details in \Cref{app:para_table_gen}).

\noindent \textbf{Data Splits} \, \,
After conducting the filtering process using the \textbf{ECs}, we selected 1,389 articles from the Pew Research corpus, 55 stories from Tableau story dashboards, and 5 stories from GapMinder, and split them into training and test sets as shown in Table~\ref{tab:train-test}. To create the test set from the Pew corpus, we selected the articles that also appear in the Chart-to-Text \cite{kantharaj-etal-2022-chart} corpus, as their chart-summary pairs were identified by human annotators to ensure the quality of the test set. For the Pew training set, we used GPT-4 model-generated annotations as explained earlier.

\subsection{Features of \model}
We analyze our corpus statistics to highlight the key features of \model. More details of the corpus analysis are included in \Cref{app:corp_ana}.

\noindent\textbf{Diversity:} 
Our benchmark contains stories covering a wide range of topics, from `Politics \& Policy' to `International Affairs,' `Education,' and `Economy'  (\Cref{fig:train_topic_dist}, and \Cref{fig:test_topic_dist}). Topics in GapMinder and Tableau are more evenly distributed while Pew is dominated by `Politics \& Policy' (57.24\%).
The corpus also includes a diverse range of chart types such as bars, lines, pies, and scatter plots (\Cref{tab:chart-type}), with bar charts being the most common (78.98\%), followed by line charts (13.40\%). 
\begin{table}[t]
\centering
\caption{DataNarrative dataset statistics. Here, `V.' denotes `Verb', `T.' denotes `Token', and `rep.' denotes `repetition'. 
}
\label{tab:text-data-stats}
\resizebox{\columnwidth}{!}{%
\begin{tabular}{lcc|cc|cc}
\multirow{2}{*}{} & \multicolumn{2}{c}{\textbf{Pew}} & \multicolumn{2}{c}{\textbf{Tableau}} & \multicolumn{2}{c}{\textbf{GapMinder}} \\\cmidrule{1-7}
                   Statistics    & Train & Test & Train & Test & Train & Test \\\cmidrule{1-7}
Avg. length of Stories & 1804  & 2865 & 837   & 1009 & -     & 707  \\
Avg. \# of Tokens       & 353   & 561  & 159   & 194  & -     & 146  \\
Avg. \# of Paragraphs   & 4     & 5    & 5     & 4    & -     & 8 \\\hdashline
Avg. V. : T. ratio (↑) & 0.51  & 0.46 & 0.64   & 0.63 & -     & 0.63  \\
Avg. \# of unique V. (↑)       & 14   & 23  & 5   & 11  & -     & 5  \\
Avg. \% of diverse V. (↑)  & 44     & 47    & 25     & 30    & -     & 39 \\
\% of Intra 3-gram rep. (↓)    & 18.38     & 17.94    & 12.79     & 14.24    & -     & 11.30 \\
\% of Inter 3-gram rep. (↓)   & 14.84     & 11.28    & 0.64     & 0.45    & -     & 2.45 \\\Xhline{1pt}
\end{tabular}%
}
\end{table} 

\noindent\textbf{Long, multimodal outputs:} Unlike existing chart domain benchmarks that produce short summaries~\cite{kantharaj-etal-2022-chart} or answers \cite{masry-etal-2022-chartqa} related to charts, \model\ have stories with multiple text paragraphs (\Cref{tab:text-data-stats}), suggesting the open-ended nature of the task. Among them, Pew stories tend to be longer with an average story length of 2334.5 characters and 457 average tokens. Each story contains 4.5 charts and corresponding paragraphs on average, demonstrating the need for planning a narrative structure that has a multimodal output covering several visualizations and related texts. 

\noindent\textbf{Semantically rich stories:} To assess semantic richness, we analyzed Vocab: Token Ratio, unique verbs, diverse verbs per story, and intra/inter-story trigram repetitions, common metrics for measuring content originality and diversity in story corpus \cite{goldfarb-tarrant-etal-2020-content}. As shown in \Cref{tab:text-data-stats}, the Tableau corpus has the highest verb-to-token ratio (0.63), while the Pew has the most unique verbs (18.5) and the highest percentage of diverse verbs (45.5\%), indicating high semantic richness. Trigram repetition is also higher in Pew, likely due to the greater length of Pew stories. 

\begin{figure}[t!]
     \centering
        \includegraphics[width=\columnwidth]{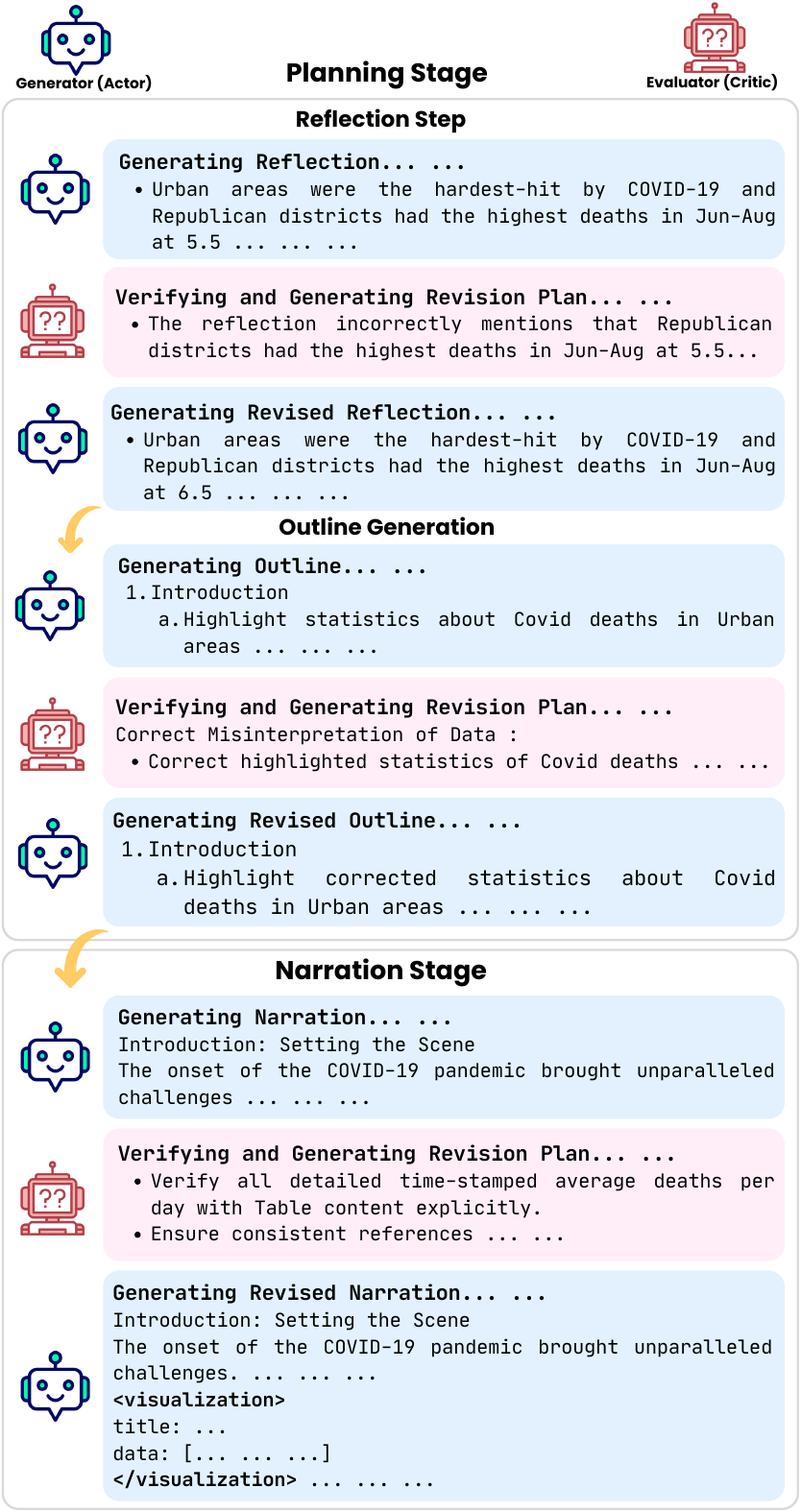}
         \caption{
         An overview of the proposed LLM-Agent framework for data story generation. 
          }
             
    \label{fig:agentic_framework}   
\end{figure}

\section{Methodology}
\label{sec:method}
\subsection{Overall Framework}
\noindent\textbf{Task Formulation:} Given one or more data table(s) and associated titles $D$, a user intent $I$ representing the main theme of the story, and additional guidelines $G$ as inputs, the expected output is a coherent data story $S$ consisting of multiple textual paragraphs and corresponding visualization specifications (e.g., chart type, x-axis/y-axis values, x-axis/y-axis labels, etc.). 
These visualization specifications are later utilized to generate visualizations based on the relevant data tables. Here, the user intent $I$ refers to the main idea or message that the author aims to convey, enabling them to achieve their communicative goal. In our corpus, we select report/story titles as user intents. 

To this end, our goal is to develop a novel multi-agent-based approach to effectively generate 
the narration of a data story. To achieve this, we propose a system that uses two LLM agents -- a Generator (Actor) and an Evaluator (Critic) -- to mimic the human process of data story generation. This process includes a planning step that involves understanding the data (reflection), creating an outline (outline generation), and the story generation step that involves narrating the story (narration), with each step being verified and revised. We introduce a pipeline approach where the response from one LLM agent serves as the context for the next agent in the sequence. In each of the stages, the generator LLM first produces an initial version of the content, which is then assessed by the critic agent based on some fixed criteria; the generator then makes a revision based on the assessment feedback (\cref{fig:agentic_framework}).

\subsection{Planning Stage}
Planning is crucial for all types of storytelling, particularly when it comes to data storytelling. The planning stage is divided into two intermediary steps: \textit{(i)} Reflection, and \textit{(ii)} Outline Generation.

\noindent{$\bullet$} \, \textbf{Reflection} \, \,
The goal of this stage is to understand and create a comprehensive description of the data presented in the data tables. 
First, the Generator Agent identifies and presents the most impactful insights, focusing on critical trends, notable patterns, and outliers that influence the overall narrative. The agent assesses the relevance, implications, and significance of the data points to determine their importance and explains the interconnections between different attributes of the data. After generating an initial reflection, the Evaluator Agent is called to verify the generation based on the data tables and asked to prepare a revision plan if necessary. At the time of verification, the Evaluator Agent cross-matches the data description with the data tables and identifies any inconsistencies and factual inaccuracies in the data description. If it determines a revision is needed, then the Generator Agent is called again to revise the initial reflection based on the revision plan. We present the prompts used at this stage in \Cref{fig:init_refl} - \ref{fig:rev_refl} in the Appendix.
The whole process can be summarized as follows: 
\begin{tcolorbox}[colback=white, colframe=black, float=h!, boxrule=0.5pt, left=1mm, right=1mm, top=1mm, bottom=1mm]
\small{\textbf{Input:} Data tables with titles $(D)$, and Additional Guidelines $(G)$.\\
\textbf{Process:}\\
(a) The Generator Agent generates initial reflections $(R_\text{init})$ in bullet points.\\
(b) Verification: The Evaluator Agent reviews the reflection, producing a revision plan $(R_\text{rvp})$ if necessary.\\
(c) Revision: The reflection is revised by the Generator Agent based on $(R_\text{rvp})$, resulting in final reflection $(R_\text{f})$.}
\end{tcolorbox}

\noindent{$\bullet$} \, \textbf{Outline Generation} \, \, Once the `reflection’ is generated, the next step in the Planning stage is outlining the data story. 
In this step, the Generator Agent constructs an outline following a linear narrative structure \cite{riche2018datadriven, segel2010narrative}, consisting of a beginning, middle, and end, to ensure a coherent flow of the story. 
It also breaks down each major point 
into smaller sub-points, highlighting specific aspects of the data such as key figures, patterns, notable exceptions, and comparisons over time and including simple visualization specifications to enhance the narrative. Additionally, the user provides an `intention' that depicts the overarching theme of the data story, and the agent is instructed to ensure that the theme is consistently emphasized throughout the outline. After generating an initial outline, the Evaluator Agent is deployed to verify the generation based on the data tables and the reflection and asked to prepare a revision plan if necessary. The agent evaluates the initial outline in two aspects, \textit{(a)} whether the insights, trends, or outliers included in the initial outline are consistent with the data presented in the tables or not, and \textit{(b)} whether the outline is coherent with the `intention' or not. If it determines a revision is needed, then the Generator Agent is called again to revise the initially generated outline accordingly.  
We present the prompts used at this stage in 
\Cref{fig:init_outl} - \ref{fig:rev_outl}.
The whole process is summarized as follows: 
\begin{tcolorbox}[colback=white, colframe=black, float=h!, boxrule=0.5pt, left=1mm, right=1mm, top=1mm, bottom=1mm]
\small{\textbf{Input:} Final reflection $(R_\text{f})$ from the previous step, data tables with titles $(D)$, and user intention $(I)$.\\
\textbf{Process:}\\
(a) The Generator Agent generates an initial outline $(O_\text{init})$ following the narrative structure.\\
(b) Verification: The Evaluator Agent reviews the outline, producing a revision plan $(O_\text{rvp})$ if necessary.\\
(c) Revision: The outline is revised based on $(O_\text{rvp})$, resulting in the final outline $(O_\text{f})$.}
\end{tcolorbox}

\subsection{Narration Stage}
The final stage of the framework is the Narration stage. The aim of this step is to generate the actual narrative text and associated visualizations. The goal is to generate a coherent data story that adheres to the narrative structure and user intention.  
The agent is also instructed to emphasize key statistics essential to understanding the theme, presenting them in a way that balances technical precision with accessibility thereby ensuring the story is approachable for both non-specialists and experts. Additionally, the agent is instructed to outline detailed specifications for visualizations, including chart titles, types (e.g., line, bar, pie, scatter plot), and axis data, where required by the outline.  
After the initial narration is generated, the Evaluator Agent assesses it to confirm its alignment with the input outline. The agent also verifies that the insights, trends, and patterns discussed are substantiated by the data tables and that the visualization specifications are factually correct. Finally, if revisions are necessary, the agent produces a revision plan. The Generator Agent then uses this plan to further refine the narration. We present the prompts used at this stage in \Cref{fig:init_narr} - \ref{fig:rev_narr}. In summary:
\begin{tcolorbox}[colback=white, colframe=black, float=h!, boxrule=0.5pt, left=1mm, right=1mm, top=1mm, bottom=1mm]
\small{\textbf{Input:} Final outline $(O_\text{f})$, data tables with titles $(D)$, and user intention $(I)$.\\
\textbf{Process:}\\
(a) The Generator Agent generates the initial narration $(N_\text{init})$, incorporating relevant story texts and vis-specs.\\
(b) Verification: The Evaluator Agent reviews the narration for factual accuracy and consistency, producing a revision plan $(N_\text{rvp})$ if necessary.\\
(c) Revision: Finally, the narration is revised based on $(N_\text{rvp})$, resulting in the final narration $(N_\text{f})$.} 
\end{tcolorbox}

In each step of the framework, the LLMs are employed three times: twice for generation and once for critique. With three steps, this totals nine LLM calls. We summarize the overall working principle of the proposed agentic framework in the algorithm provided in the \Cref{app:llmagent}.

\section{Evaluation}

\subsection{Evaluation Methods}
\label{subsec:eval_methods}
We employed GPT-4o \cite{OpenAI2024}, LLaMA-3-8b-instruct, and LLaMA-3-70b-instruct \cite{MetaAI2024} models as the Generator and Evaluator Agents for story generation. GPT-4o was chosen for its exceptional performance across various NLP downstream tasks \cite{OpenAI2024}. Additionally, we utilized the leading open-source model LLaMA-3-70b-instruct and the smaller-scale option LLaMA-3-8b-instruct \cite{chiang2024chatbot}. To generate the stories, we used the data tables from our test set which has 339 stories. To assess the efficacy of the agentic framework for story generation, we used two rigorous evaluation methods: \textit{(i)} automatic evaluation using Gemini-1.5-pro 
~\cite{geminiteam2024gemini} as an LLM-judge and \textit{(ii)} human evaluation. 

\subsection{Automatic Evaluation} 
\noindent\textbf{Method} \, Previous studies have found that reference-based evaluation metrics like the BLEU score often do not align with the attributes of text quality as perceived by humans \cite{smith-etal-2016-climbing, liu2023geval}. In addition, given the inherently objective nature of the story generation task, especially in data story generation, we established comprehensive methods for both automatic and human evaluations. Following the work of \citet{zheng2023judging} and \citet{yuan2024selfrewarding}, we implemented an automatic evaluation method i.e., pairwise comparison of the stories generated by the agentic framework versus direct prompting. The evaluation criteria included `Informativeness', `Clarity and Coherence', `Visualization Quality', `Narrative Quality', and `Factual Correctness'. 

\noindent\textbf{Results} \, As illustrated in \Cref{tab:agentvsnonagent}, the agentic framework significantly outperformed the direct approach, as demonstrated by GPT-4o, which attained an average win rate of 75.93\% across three test sets, compared to the direct approach's 23.47\%, highlighting a substantial difference of 52.46\%.
Similarly, LLaMA-3-70b-instruct using the agentic approach attained an average win rate of 58.7\%, while the direct approach only achieved 39.82\%. These results indicate a clear preference by the LLM judge (Gemini-1.5-pro-001 in our case) for stories generated with the agentic approach over direct prompting. 
However, the LLaMA-3-8b-instruct model demonstrated balanced performance with our agentic approach outperforming its counterpart in only 40.59\% of cases.
This outcome may be attributed to its relatively smaller size, 
and its limited 8k context length. 
These factors indicate that there is still potential for improvement through task-specific fine-tuning. Overall, these findings underscore the superior efficacy of the LLM-agent framework in producing coherent data stories.

\begin{figure*}[t!]
     \centering
        \includegraphics[width=\textwidth]{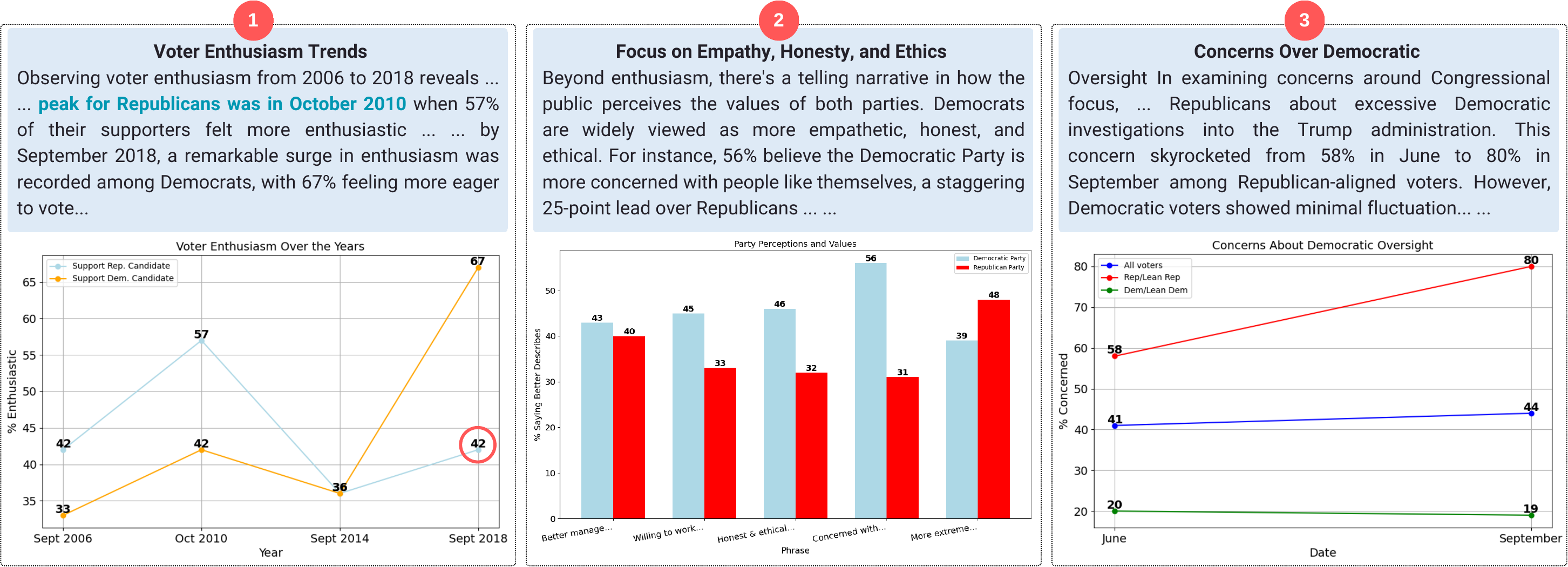}
         \caption{
         An example of a GPT-4o-generated story using the agentic framework: The text in \textcolor{moonstoneblue}{\textbf{Blue}} color denotes hallucinated fact, while the \textcolor{red}{\textbf{red circled}} value is factually incorrect according to `\textit{Table\_0}' of \Cref{fig:fact_hall_gpt4o_table}.       
          }
    \label{fig:fact_hall_gpt4o}
\end{figure*}

\begin{table}[t]
\centering
\caption{An overview of the results from automatic evaluation with pairwise comparison.}
\label{tab:agentvsnonagent}
\resizebox{0.9\columnwidth}{!}{%
\begin{tabular}{lccc}
\midrule
\textbf{Model} &
  \begin{tabular}[c]{@{}c@{}}Agentic\\ Win (\%)\end{tabular} &
  \begin{tabular}[c]{@{}c@{}}Direct\\ Win (\%)\end{tabular} &
  \begin{tabular}[c]{@{}c@{}}Tie\\ (\%)\end{tabular} \\\midrule
GPT-4o               & \textbf{78.17}       & 20.05       & 1.78      \\
  LLaMA-3-70b-instruct & \textbf{58.70}       & 39.82       & 1.48    \\ LLaMA-3-8b-instruct  & 41.59       & \textbf{54.57}       & 3.84  
 \\\Xhline{1pt}    
\end{tabular}%
}
\end{table}
\subsection{Human Evaluation}
\noindent\textbf{Method} \, For human evaluation, in line with similar research in story generation \cite{wang-etal-2023-improving-pacing, yang-etal-2023-doc}, we assess the stories produced by the LLMs using various subjective metrics. These metrics include `Informativeness', `Clarity and Coherence', `Visualization Quality', `Narrative Quality', and `Factual Correctness'. 
We conducted a human evaluation on 100 story samples generated by the top-performing model (GPT-4o). For each sample, two annotators performed a pairwise comparison between the two versions, one generated by the agentic framework and the other one by the direct prompting method, and the agreement between them for these comparisons was 85.0\%.

\begin{table}[t]
\centering
\caption{Human evaluation results of the story generation setup: GPT-4o (Agentic) vs. GPT-4o (Direct)
}
\label{tab:agentvsnonagenthum}
\resizebox{\columnwidth}{!}{%
\begin{tabular}{lc|c|c|c}
 & \multicolumn{4}{c}{\textbf{GPT-4o (Agentic vs. Direct)}} \\ \Xhline{1pt}
\textbf{Metrics} & \begin{tabular}[c]{@{}c@{}}Agentic \\ Win (\%)\end{tabular} & \begin{tabular}[c]{@{}c@{}}Direct \\ Win (\%)\end{tabular} & \begin{tabular}[c]{@{}c@{}}Tie \\ (\%)\end{tabular} & \begin{tabular}[c]{@{}c@{}}$p$-value \\ (sign test)\end{tabular} \\ \cmidrule{1-5}
Informativeness & \textbf{74} & 11 & 15 & $1.29 \mathrm{e}{-12}$  \\ 
Clarity and  Coherence & \textbf{73} & 11 & 16 & $2.25 \mathrm{e}{-12}$ \\ 
Visualization Quality & \textbf{59} & 15 & 26 & $2.55 \mathrm{e}{-07}$  \\ 
Narrative Quality & \textbf{75} & 12 & 13 & $2.71 \mathrm{e}{-12}$ \\ 
Factual Correctness & \textbf{75} & 11 & 14 & $7.37 \mathrm{e}{-13}$ \\ \Xhline{1pt}
\end{tabular}%
}
\end{table}

\noindent\textbf{Results} \, The results from \Cref{tab:agentvsnonagenthum} indicate that the stories generated by the agentic approach are of significantly higher quality compared to those produced by the non-agentic version. This is demonstrated by an impressive average win rate of 71.2\% across all five evaluation criteria. Furthermore, we compared the human-evaluated stories with our automatic evaluation and found that our human annotators agreed with the LLM judge in 67.0\% of the cases, suggesting that human annotators' scores are roughly consistent with the LLM judge.

\subsection{Ablation Studies}
To assess the efficacy of the agentic approach, we perform ablation experiments on a randomly selected subset of 100 stories and evaluate them automatically by the LLM judge (Gemini-1.5-pro-001). These experiments focused on excluding different steps (see \Cref{tab:ablation_strat}) and comparing the generated stories with those produced by the agentic approach.

From \noindent\Cref{tab:ablation_results}, we observe that The most significant decline occurred when all steps, especially when the Planning stage (Reflection and Outline Generation), were skipped (79\% loss). Skipping either the Reflection or Outline Generation step also led to a decline in performance, though less severe, with a 64\% loss in both cases. This demonstrates that the agentic framework's performance is roughly twice as effective as other approaches, underscoring its importance and value. Finally, omitting the verification step resulted in a 73\% loss, compared to a 22\% case of win, emphasizing the crucial role of the `Critic' agent in the framework. 

\begin{table}[t]
\centering
\caption{Ablation Strategy. Here, `Refl', `Out.', `Narr.', and `Ver' denotes `Reflection', `Outline', `Narration', and `Verification' respectively}
\label{tab:ablation_strat}
\resizebox{\columnwidth}{!}{%
\begin{tabular}{cccc|cc}
  \multicolumn{4}{c}{\textbf{Planning Stage}} &
  \multicolumn{2}{c}{\textbf{Narration Stage}} \\\midrule
  Refl. &
  Refl. ver. &
  Out. Gen. &
  Out. ver. &
  Narr. &
  Narr. ver. \\ \midrule
  {\color[HTML]{32CB00} \cmark} &
  {\color[HTML]{32CB00} \cmark} &
  {\color[HTML]{32CB00} \cmark} &
  {\color[HTML]{32CB00} \cmark} &
  {\color[HTML]{32CB00} \cmark} &
  {\color[HTML]{32CB00} \cmark} \\
  {\color[HTML]{FD6864} \xmark} &
  {\color[HTML]{FD6864} \xmark} &
  {\color[HTML]{32CB00} \cmark} &
  {\color[HTML]{32CB00} \cmark} &
  {\color[HTML]{32CB00} \cmark} &
  {\color[HTML]{32CB00} \cmark} \\
  {\color[HTML]{32CB00} \cmark} &
  {\color[HTML]{32CB00} \cmark} &
  {\color[HTML]{FD6864} \xmark} &
  {\color[HTML]{FD6864} \xmark} &
  {\color[HTML]{32CB00} \cmark} &
  {\color[HTML]{32CB00} \cmark} \\
  {\color[HTML]{FD6864} \xmark} &
  {\color[HTML]{FD6864} \xmark} &
  {\color[HTML]{FD6864} \xmark} &
  {\color[HTML]{FD6864} \xmark} &
  {\color[HTML]{32CB00} \cmark} &
  {\color[HTML]{32CB00} \cmark} \\
  {\color[HTML]{32CB00} \cmark} &
  {\color[HTML]{FD6864} \xmark} &
  {\color[HTML]{32CB00} \cmark} &
  {\color[HTML]{FD6864} \xmark} &
  {\color[HTML]{32CB00} \cmark} &
  {\color[HTML]{FD6864} \xmark} \\\Xhline{1pt}
\end{tabular}%
}
\end{table}
\begin{table}[t]
\centering
\caption{The results from our ablation experiment in four different setups. We report the `Loss', `Win', and `Tie' of different setups against the Agentic framework.
}
\label{tab:ablation_results}
\resizebox{\columnwidth}{!}{%
\begin{tabular}{lccc}
\midrule
\textbf{Strategy}                                                        & Loss (\%) & Win (\%) & Tie (\%) \\\midrule
w/o `Reflection'                                                         & 64\%        & 35\%       & 1\%        \\
w/o `Outline'                                                            & 64\%        & 32\%       & 4\%        \\
w/o `Reflection' and `Outline' & 79\%        & 18\%       & 3\%        \\
w/o `Verification'                                                       & 73\%       & 22\%       & 5\%   \\\Xhline{1pt}    
\end{tabular}%
}
\end{table}

\subsection{Error Analysis and Challenges}
We manually analyzed 100 sample data stories generated by the agentic framework to understand the key challenges in addressing our new task.

\noindent \textbf{Factual errors:}
Despite the verification steps at each stage, factual errors sometimes occur during the narration phase. For instance, the red circle in slide (1) of \Cref{fig:fact_hall_gpt4o} highlights a factual error where the actual value is 59\% instead of 42\%, as per `\textit{Table\_0}' of \Cref{fig:fact_hall_gpt4o_table}.

\noindent \textbf{Hallucination errors} \,\,
Although hallucinating facts is a rare occurrence in the GPT4o-generated stories using the agentic approach, some cases appear where the model is prone to hallucinating facts. For example in \Cref{fig:fact_hall_gpt4o},  the model mentions that `the peak of Republican enthusiasm was in `October 2010', whereas according to `\textit{Table\_0}' of \Cref{fig:fact_hall_gpt4o_table} it was `September 2018' at 59\%.

\noindent \textbf{Ambiguous visualization specifications} \,\,
In some cases,  the model generates ambiguous chart specifications such as `side-by-side bar chart,' `multi-dimensional infographic,' `summary chart,' or `combined' as chart types. Such ambiguous specifications make it difficult to render charts correctly, illustrating the limitations of existing models in generating multimodal outputs with charts.

\noindent \textbf{Lack of coherence and verbosity issue} \,\,
A key challenge faced by the open-source LLaMA-3 models is maintaining a coherent narrative structure, particularly when using the agentic approach which tends to produce more verbose text. On average, the length of stories generated by the LLaMA-3-8b-instruct model is approximately 610 tokens, while those generated using the non-agentic approach contain about 500 tokens. \Cref{fig:coher_8b} shows that 
despite the story's theme being the `\textit{EU's response to COVID-19},' the third slide features unrelated statistics, and the fourth slide repeats text from the third.
This highlights the limitations of relatively smaller open-source LLMs (8B) in producing long, multimodal stories with complex narratives. 

\section{Conclusion and Future Work}
We present \model, a new benchmark for multimodal data story generation that combines text generation, data analysis, and information visualization. Our benchmark includes 1,449 diverse data stories with open-ended multimodal outputs, each featuring various charts and related texts. We then propose an LLM-agent-based story generation framework that mimics the human process of creating data stories by using a generator and an evaluator agent. Our experiments show that this framework generally outperforms the direct method in both automatic and human evaluations.

The study also highlights unique challenges in multimodal long-form data story generation, such as the difficulty of building open-source models that generate long, coherent stories with rich narratives. To address this, we release a training corpus for the community to explore fine-tuning open-source models for this task. Additionally, our agentic framework can serve as a foundation for human-in-the-loop co-authoring of data stories with LLMs, where humans act as critics, collaborating and co-editing with the LLM to create coherent and informative stories. We hope our research inspires further work in multimodal data storytelling.

\section*{Acknowledgement}
The authors would like to thank the anonymous reviewers for their helpful comments. The authors would also like to thank Mizanur Rahman for his valuable contributions to the human evaluation process. This research was supported by the Natural Sciences and Engineering Research Council (NSERC), Canada, Canada Foundation for Innovation, and the CIRC  grant on Inclusive and Accessible Data Visualizations and Analytics.

\section*{Limitations}
Despite the fact that the proposed agentic framework is capable of producing coherent and informative data stories, there are instances where the model may generate factually inaccurate statements within the text. Furthermore, in certain rare cases, the visualization specifications might be sufficient to create a chart image but may still lack critical information. Furthermore, because of the expense associated with API access, we were unable to assess other state-of-the-art proprietary LLMs similar to GPT-4o, such as Claude-3 \cite{Claude}. Due to resource constraints, we were unable to fine-tune an open-source model within the limited time available. However, we plan to release a fine-tuned model as part of our future research. Additionally, we will make the training corpus available to the community to facilitate further exploration of fine-tuning open-source models for this task.

\section*{Ethics Statement}

At the time of the dataset collection process, we carefully considered various ethical aspects. The three sources of our data story corpus (Pew Research Center \cite{pewresearch}, Tableau Public \cite{tableaupublic}, and GapMinder \cite{gapminder}) approve publication rights for academic utilization of their content. We plan to make the whole corpus and all the collected metadata publicly available.  

To ensure our chart images are free of harmful content, we utilized Google search, benefiting from its rigorous content policies\footnote{\href{https://blog.google/products/search/when-and-why-we-remove-content-google-search-results/}{https://blog.google/products/search/when-and-why-we-remove-content-google-search-results/}}. Moreover, during the data extraction process, the chart images were analyzed using the Gemini API, which is specifically designed to filter out unsafe content\footnote{\href{https://ai.google.dev/docs/safety\_setting\_gemini}{https://ai.google.dev/docs/safety\_setting\_gemini}}, thereby ensuring an additional degree of certainty concerning the appropriateness of the content included in our dataset.

The human evaluation was conducted by the authors and their collaborators associated with this research. Since the primary aim was to assess the models' capabilities, effectiveness, and limitations in generating stories across various experimental conditions, the evaluation by the authors does not introduce any ethical concerns or unwanted biases. The instructions given to the human evaluators are provided in \Cref{fig:human_eval_instruction}. There were no paid participants in the human evaluation study. For the human evaluation study, we selected two human evaluators. The first evaluator has more than three years of industry experience (also has a graduate degree in computer science) in data science and information visualization. The second evaluator comes from an academic background (and has an undergraduate degree in computer science) and has one year of experience in information visualization. Additionally, since the evaluators were volunteers, there were no paid participants in the human evaluation study.
Lastly, the evaluation did not involve any information that could be used to identify individuals.

\bibliographystyle{acl_natbib}
\bibliography{ms}

\newpage
\appendix
\begin{appendices}
% \section{Appendix}
\section{Dataset Construction Process}
\label{app:dataset}
In this section, we provide further detail on our dataset curation process. 

\subsection{Data Sources}
The corpus for \model \, consists of stories collected from three different platforms: Pew Research \citet{pewresearch}, Tableau Public Data Story \citet{tableaupublic}, and Gapminder \cite{gapminder}. Pew Research releases articles based on data that focus on social issues, public opinion, and demographic trends. These articles frequently include various charts and are complemented by high-quality descriptions from professional editors. Gapminder is a Swedish foundation dedicated to fighting misconceptions about global development by promoting a fact-based worldview. They provide interactive data visualization tools and publish educational resources, such as data stories, and interactive visualizations that emphasize global trends in health, wealth, and other development indicators. On the other hand, Tableau Public Story, a feature of Tableau Public, is a platform that enables users to create interactive presentations through a series of data visualizations. It makes data stories publicly accessible, covering a wide range of topics including economy, social issues, and international affairs. Therefore, the corpus benefits from this diversity by providing stories with varying topics, styles, and themes. 

\subsection{Raw Data Collection}
% \noindent{$\bullet$} \, \textbf{Pew} \, \, 
To assemble the Pew corpus, we created a web crawling script that initially stores research topics and their corresponding URLs. This script systematically processes the HTML elements from these URLs to collect all links, categorizing them under general topics while excluding irrelevant ones like ``Methodological Research'' and ``Full topic list'' that do not link to any meaningful article webpage. Subsequently, another script is employed to visit all the article pages for each topic, extracting and parsing HTML content to gather various data such as article texts, titles, and image links. These image links are then filtered by specific criteria (e.g., `jpg', `jpeg', `SVG', or `png' formats) to ensure data integrity, eliminating duplicates. A secondary script downloads these images in `PNG' format. We gathered articles from the Pew Research website until March 14, 2024, resulting in 4532 articles across 18 topics. Additionally, we collected metadata related to the images, including captions and alt-texts. 

\begin{figure*}[t]
     \centering
        % \vspace{mm}
        \includegraphics[width=\textwidth]{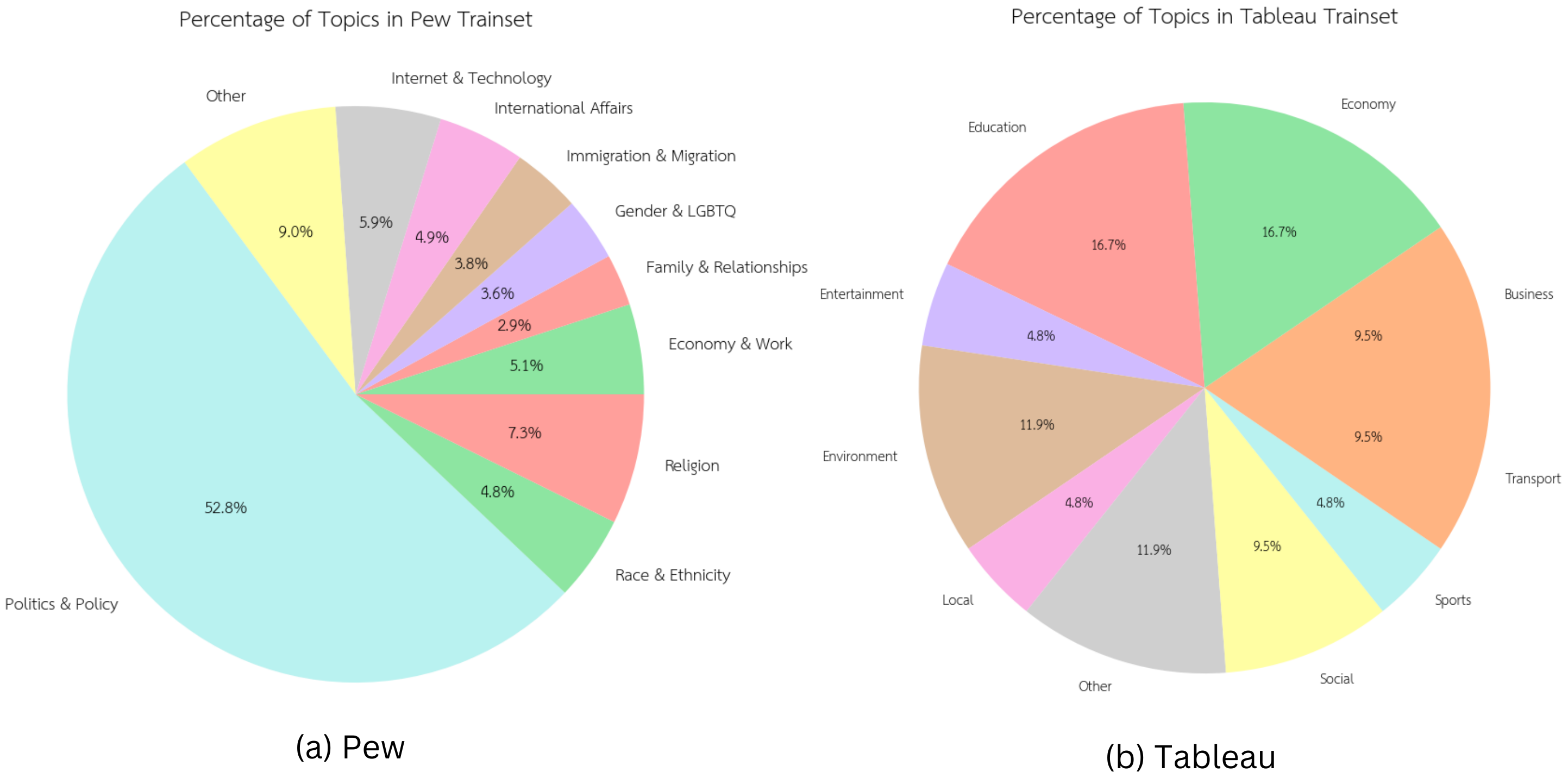}%emnlp2020-templates/Figures/Pipeline.png}
        % \vspace{-3mm}
         \caption{
         The figure demonstrates the distribution of Story Topics in the Train set.
         % \vspace{-3mm}
          }
    \label{fig:train_topic_dist}
\end{figure*}

\begin{figure*}[t]
     \centering
        % \vspace{mm}
        \includegraphics[width=\textwidth]{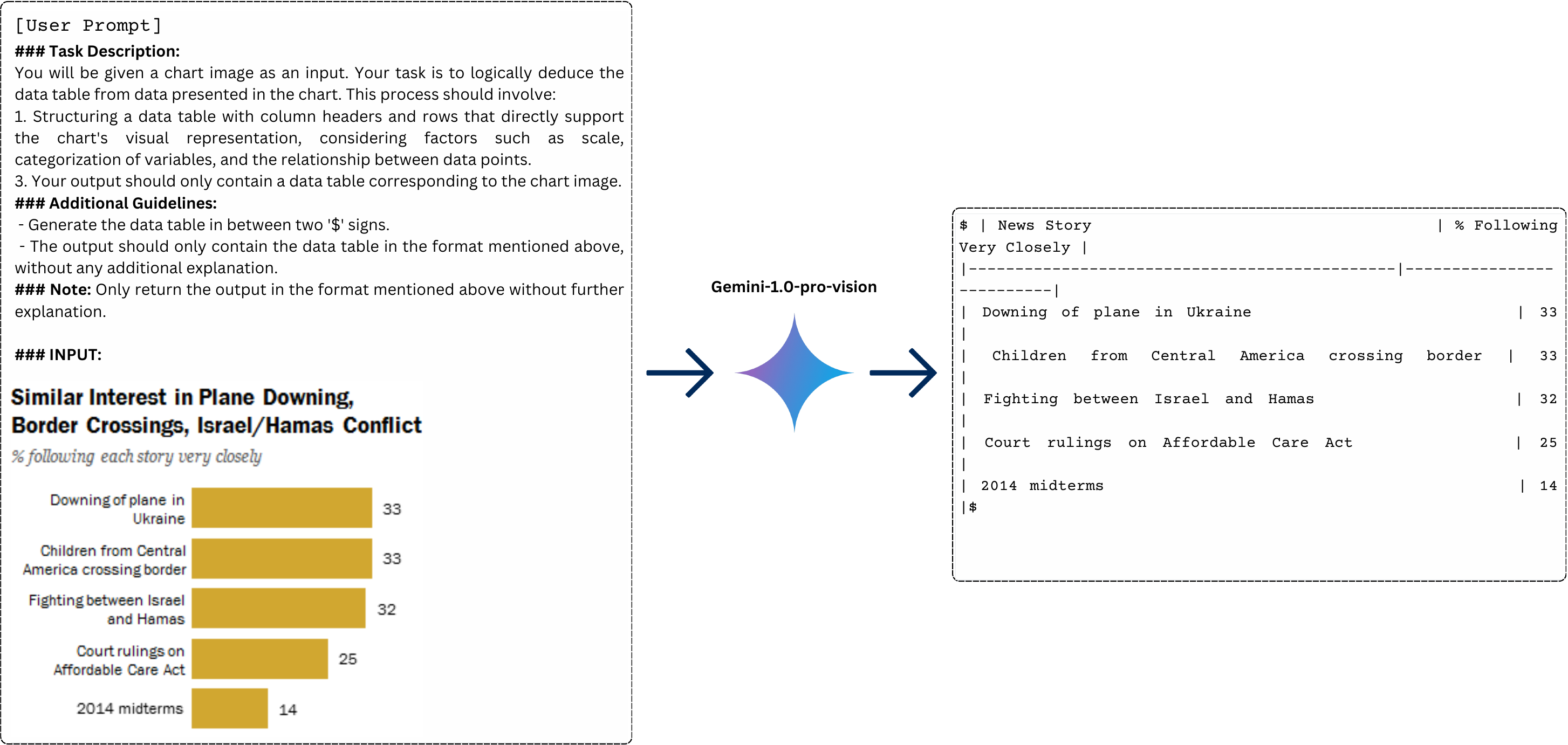}%emnlp2020-templates/Figures/Pipeline.png}
        % \vspace{-3mm}
         \caption{
         The figure presents an overview of the Chart data extraction process using the Gemini-1.0-pro-vision \cite{geminiteam2023gemini} model.
         % \vspace{-3mm}
          }
    \label{fig:chart_data_extract}
\end{figure*}

\subsection{Chart Data Extraction}
\label{app:chart_data_ext}
We utilize the multi-modal large language model (MLLM) Gemini-1.0-pro-vision \cite{geminiteam2023gemini} to extract data from chart images. In order to verify the factual correctness of the generated data tables, we conducted a small experiment using 100 chart images from the ChartQA \cite{masry-etal-2022-chartqa} corpus, where gold tables were already available, allowing for direct comparison between the gold tables and the generated tables. We performed a human evaluation of the generated data tables and found that the model correctly generated the tables in 77\% of the cases. Most errors occurred when the model either produced incomplete tables (missing one or two values or an entire row) or failed to generate any output at all. \Cref{fig:chart_data_extract} presents an overview of the chart data extraction process.

\begin{figure*}[t]
     \centering
        % \vspace{mm}
        \includegraphics[width=\textwidth]{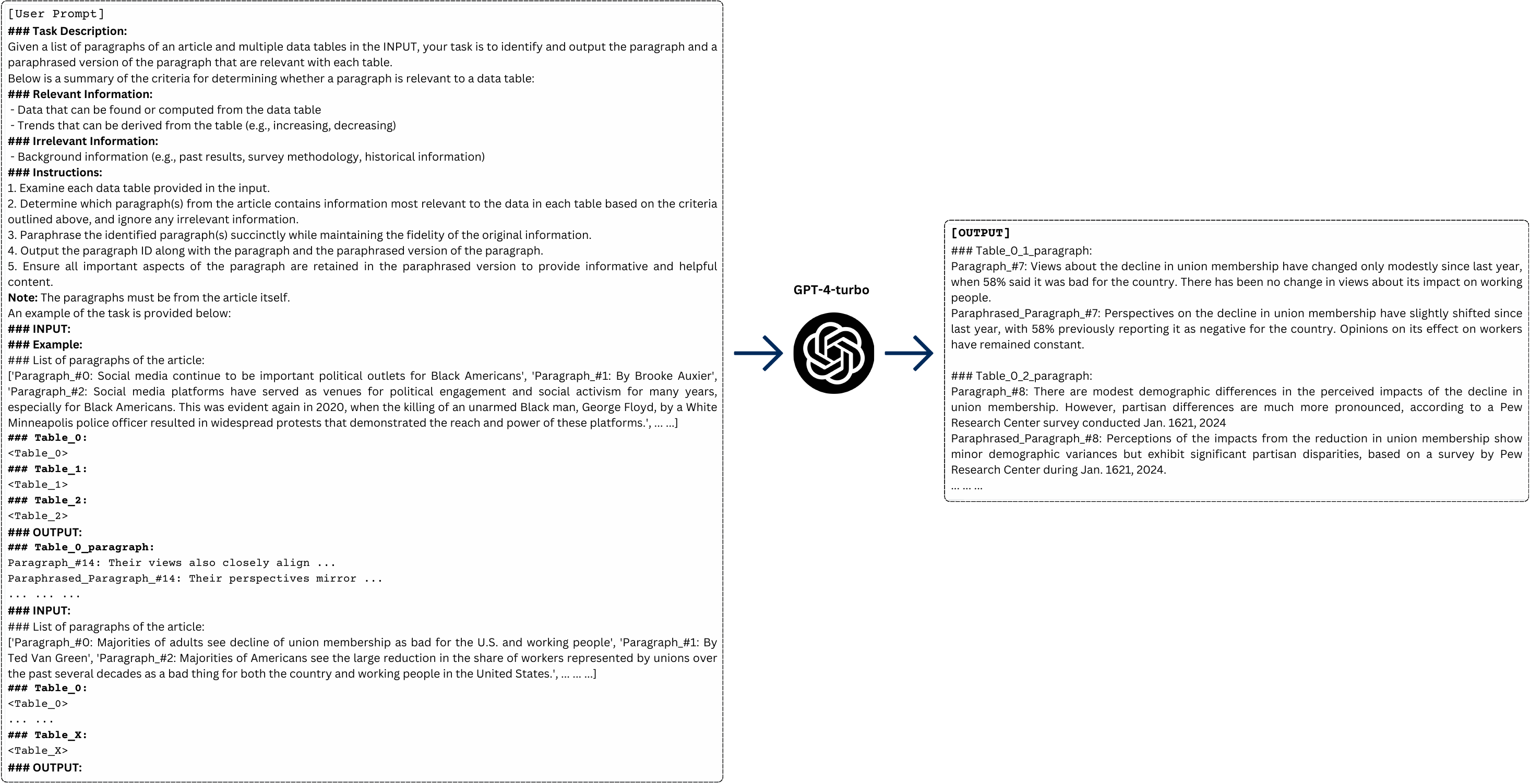}%emnlp2020-templates/Figures/Pipeline.png}
        % \vspace{-3mm}
         \caption{
         The figure presents an overview of the Paragraph table pair generation using the GPT-4-turbo \cite{gpt4turbo} model.
         % \vspace{-3mm}
          }
    \label{fig:para_table_gen}
\end{figure*}
\subsection{Chart-text pair Collection}
\label{app:para_table_gen}
As the Pew corpus is larger than the other corpora, collecting paragraphs associated with the data tables manually is labor-intensive and time-consuming. Therefore, for the Pew training set, we adopted an automatic approach using the GPT-4-turbo model \cite{gpt4turbo}. The model selected relevant paragraphs from articles based on data tables for the chart images that we extracted automatically. In addition to collecting the original paragraphs, we also generated the paraphrased version of the paragraphs using the GPT-4-turbo model as well. To evaluate the effectiveness and accuracy of this approach, we compared human-curated paragraphs from Pew articles with those selected by GPT-4-turbo. By examining 50 randomly selected samples from the Chart-to-Text corpus, we found that GPT-4-turbo accurately linked paragraphs to data tables 70\% of the time. As a result, we decided to use GPT-4-turbo-generated paragraphs for the Pew training set. To create the test set from the Pew corpus, we selected the articles and the paragraph-table pairs from each of the articles that appear in the Chart-to-Text \cite{kantharaj-etal-2022-chart} Pew corpus. \Cref{fig:para_table_gen} illustrates an overview of the chart-text collection process.
\begin{figure*}[t]
     \centering
        % \vspace{mm}
        \includegraphics[width=\textwidth]{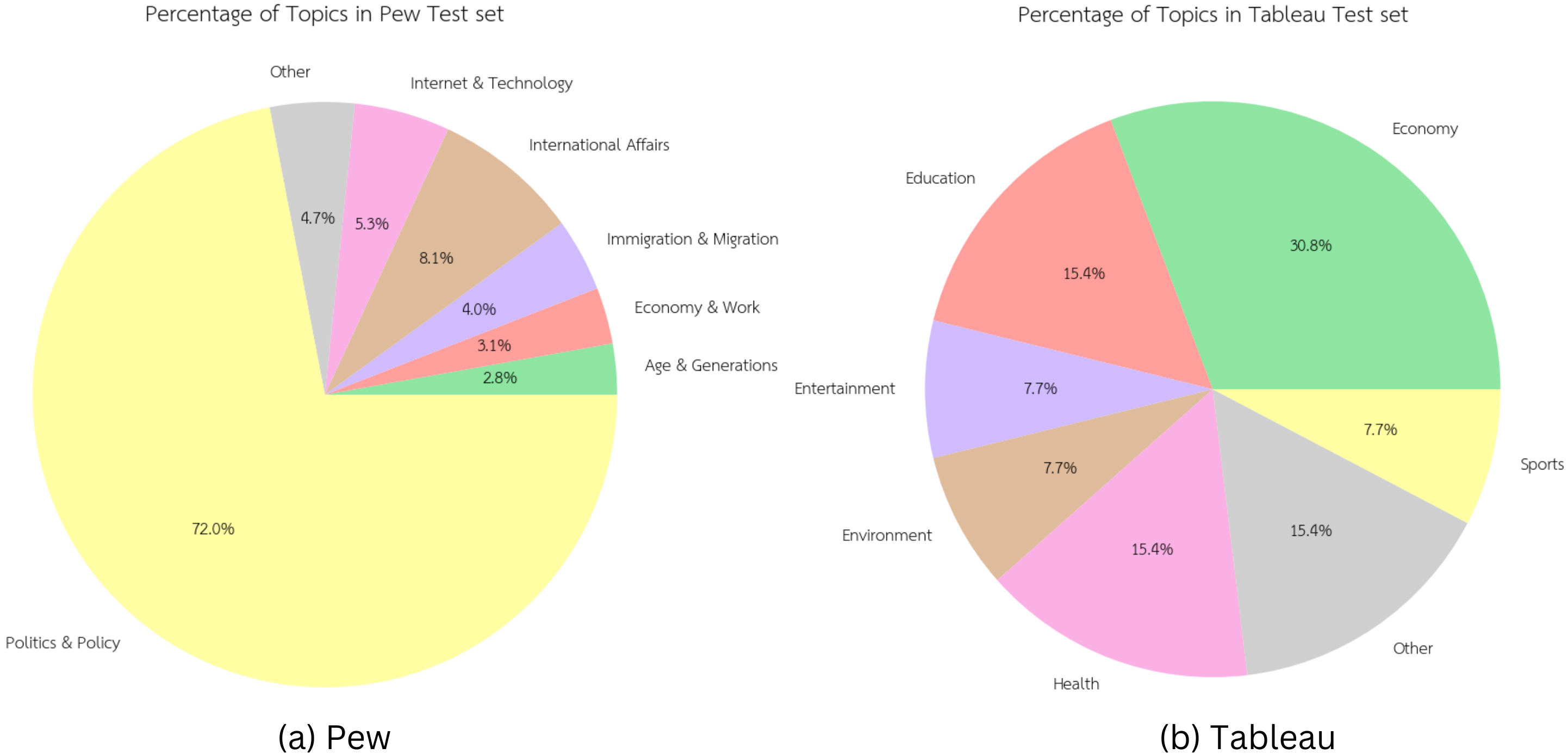}%emnlp2020-templates/Figures/Pipeline.png}
        % \vspace{-3mm}
         \caption{
         The figure demonstrates the distribution of Story Topics in the Test set.
         % \vspace{-3mm}
          }
    \label{fig:test_topic_dist}
\end{figure*} 

\subsection{Detailed Corpus Analysis}
\label{app:corp_ana}
In this section, we present a more fine-grained analysis of the proposed dataset for \model. 

\noindent{$\bullet$} \, \textbf{Pew} \,\, 
The Pew training corpus includes 1,068 stories, encompassing a total of 4,729 tables and 4,729 charts. On average, the length of these stories is 1,804 characters, consisting of an average of 353 tokens and organized into on average 4 paragraphs per story. The vocabulary-to-token ratio averages 0.51, with each story typically featuring 14 unique verbs, and 44\% of these verbs are diverse. Trigram repetition within stories stands at 18.37\%, while between stories it is 14.83\%. From \Cref{tab:chart-type} we observe that in the Pew train set, a significant majority of the charts are bar charts (both simple as well as stacked and group bar charts) (83.51\%), followed by line charts (9.16\%), and pie charts (4.04\%), etc. Regarding topic variety, 51.84\% of the stories focus on `Politics \& Policy', 7.17\% on `Religion', and 5.79\% on `Internet \& Technology', among other categories.

The Pew test corpus comprises a total number of 321 stories, with a total of 1590 tables and 1590 charts. The average length of stories in the train set is 2865 characters, the average token count is 561 and there are 5 paragraphs in each sample story on average. Additionally, the average vocabulary-to-token ratio is 0.46, with an average of 23 unique verbs per story, and 47\% of the verbs used are diverse. The intra-story trigram repetition rate is 17.94\%, while inter-story trigram repetition is 11.28\%. Similarly, \Cref{tab:chart-type} indicates that in the Pew test set, the majority of the charts are bar charts (simple, stacked, and group) at 77.79\%, followed by line charts at 17.45\%, and pie charts at 3.56\%. Regarding topic diversity, about 71.96\% of the stories are related to `Politics \& Policy', 8.09\% to `International Affairs', and 5.29\% to `Internet \& Technology'.

\begin{figure}[t]
     \centering
        % \vspace{mm}
        \includegraphics[width=\columnwidth]{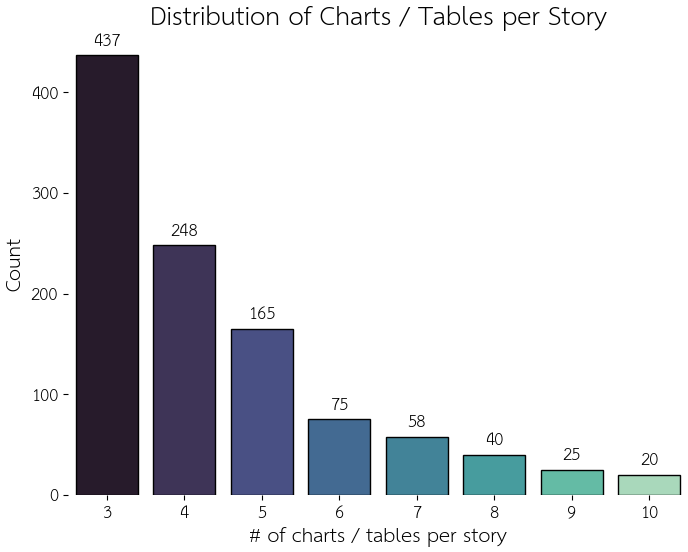}%emnlp2020-templates/Figures/Pipeline.png}
        % \vspace{-3mm}
         \caption{
         Distribution of \# of charts / tables per story (Pew Train).
         % \vspace{-3mm}
          }
    \label{fig:pew_train_ch_tb_dist}
\end{figure} 
\noindent{$\bullet$} \, \textbf{Tableau} \,\, The training corpus for Tableau consists of 42 stories with a total of 340 tables and 297 charts. Each story in the training set averages 837 characters, 159 tokens, and 5 paragraphs. The vocabulary-to-token ratio averages 0.64, and each story typically includes 5 unique verbs, with 25\% of them being diverse. The percentage of intra-story trigram repetition is 12.79\% and inter-story trigram repetition is 0.64\%. The Tableau test corpus consists of 13 stories, with 64 tables and 64 charts. From \Cref{tab:chart-type} we can see that bar charts are the most common chart type in the Tableau train set, accounting for 52.19\% of all charts. They are followed by line charts (23.23\%) and scatter plots (12.12\%). In terms of topic diversity, approximately 16.67\% of the stories are about the `Economy', followed by `Education' (16.67\%) and the `Environment' (11.9\%), among others.

In the test set, the average story length is 1009 characters, the average token count is 194, and each story contains an average of 4 paragraphs. Additionally, the vocab: token ratio is 0.63, the average number of unique verbs per story is 11, and 30\% of the verbs in a story are diverse. The percentage of intra-story trigram repetition is 14.24\%, and the percentage of inter-story trigram repetition is 44.67\%. Similarly, regarding the charts in the Tableau test set, \Cref{tab:chart-type} shows that bar charts (simple, stacked, and grouped) comprise the majority (71.88\%), followed by line charts (12.5\%) and scatter plots (9.37\%). In terms of topic diversity, approximately 30.77\% of the stories are about the `Economy', followed by `Education' (15.38\%) and the `Environment' (7.69\%), among others. 
\begin{figure}[t]
     \centering
        % \vspace{mm}
        \includegraphics[width=\columnwidth]{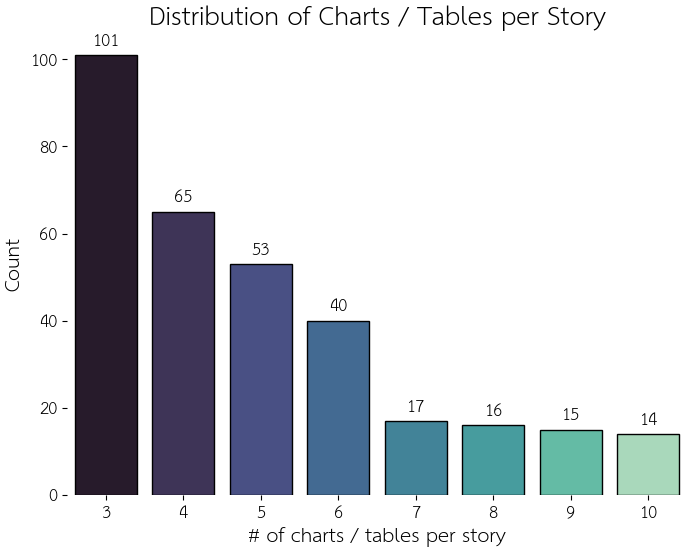}%emnlp2020-templates/Figures/Pipeline.png}
        % \vspace{-3mm}
         \caption{
         Distribution of \# of charts / tables per story (Pew Test).
         % \vspace{-3mm}
          }
    \label{fig:pew_test_ch_tb_dist}
\end{figure}

\noindent{$\bullet$} \, \textbf{Gapminder} \,\, The GapMinder test corpus consists of five stories, with a total of 42 tables and 42 charts. The average length of stories in the train set is 707 characters, and there are 8 paragraphs in each sample story on average. The average token count is 146. Additionally, the average vocab: token ratio is 0.63, the average number of unique verbs per story is 5, and there are 39\% of diverse verbs present in a story. Furthermore, the percentage of intra-story trigram repetition is 11.3\% and inter-story trigram repetition is 2.45\%. From \Cref{tab:chart-type} we observe that the Gapminder dataset mainly focuses on topics such as `World Population', `World Economy', and `Population Birthrate'. The dataset only consists of line charts (73.81\%) and bubble charts (26.19\%).

In addition, \Cref{fig:train_topic_dist} and \Cref{fig:test_topic_dist} detail the overall topic distribution in the train and test set respectively. Furthermore, \Cref{fig:pew_train_ch_tb_dist} and \Cref{fig:pew_test_ch_tb_dist} show the distributions of Charts / Tables per Story in the Pew train and test set respectively.

\section{LLM Agent Framework}
\label{app:llmagent}
We summarize the whole working process of the proposed agentic framework in the \Cref{algorithm}:
\begin{algorithm}[t!]
\SetAlgoNlRelativeSize{0}
\caption{Data Story Generation Framework}
\label{algorithm}
\SetAlgoNlRelativeSize{0}
\KwIn{Data tables with titles $D$, Additional Guidelines $G$, Intention $I$}
\KwOut{Final narration $N_f$}

$R_0 \gets \text{Generate}(D, G)$ \tcp*[r]{Generate initial reflection}
$V_R \gets \text{Verify}(D, R_0)$ \tcp*[r]{Verify reflection}
$R_f \gets \text{Revise}(R_0, V_R)$ \tcp*[r]{Revise reflection}

$O_0 \gets \text{Generate}(R_f, D, I)$ \tcp*[r]{Generate initial outline with intention}

$V_O \gets \text{Verify}(D, R_f, O_0)$ \tcp*[r]{Verify outline}
$O_f \gets \text{Revise}(O_0, V_O)$ \tcp*[r]{Revise outline}

$N_0 \gets \text{Generate}(O_f, D, I)$ \tcp*[r]{Generate initial narration with intention}

$V_N \gets \text{Verify}(D, O_f, N_0)$ \tcp*[r]{Verify narration}
$N_f \gets \text{Revise}(N_0, O_f, V_N, I)$ \tcp*[r]{Revise the narration (if necessary) and generate the final version}

\end{algorithm}

\section{Additional Results and Evaluation Details}
\label{app:add_res}
\begin{figure*}[t]
     \centering
        % \vspace{mm}
        \includegraphics[width=\textwidth]{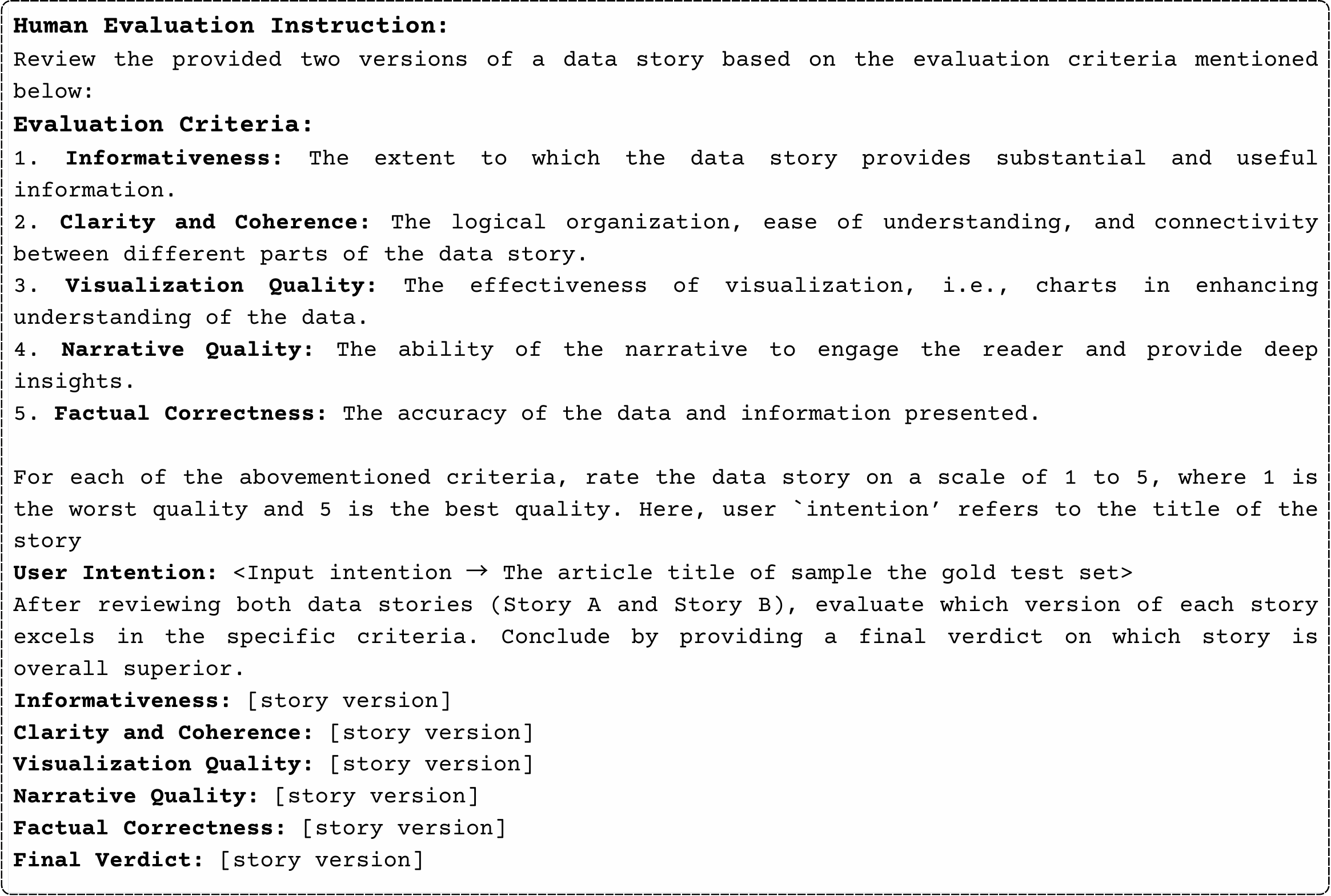}%emnlp2020-templates/Figures/Pipeline.png}
        % \vspace{-3mm}
         \caption{
         Instruction for our Human Evaluation settings.
         % \vspace{-3mm}
          }
    \label{fig:human_eval_instruction}
\end{figure*}

In this section, we detail our human evaluation approach and present a detailed result analysis (see \Cref{fig:human_eval_instruction})\\
\noindent \textbf{Human Evaluation} \,\, Our human evaluation metrics include `Informativeness', `Clarity and Coherence', `Visualization Quality', `Narrative Quality', and `Factual Correctness'. Below we present the description of the metrics:\\
\textit{(a)} Informativeness: The extent to which the data story provides substantial and useful information. \\
\textit{(b)} Clarity and Coherence: The logical organization, ease of understanding, and connectivity between different parts of the data story.\\
\textit{(c)} Visualization Quality: The effectiveness of visualization, i.e., charts in enhancing understanding of the data.\\
\textit{(d)} Narrative Quality: The ability of the narrative to engage the reader and provide deep insights.\\
\textit{(e)} Factual Correctness: The accuracy of the data and information presented.

We assessed each story using two human annotators for each evaluation criterion. For every story, we presented two versions—one generated using the Agentic framework and the other using the Direct prompting method—without disclosing which version was which. The annotators were then asked to determine which version was superior based on each criterion. In cases where the annotators disagreed, we considered the result as a tie. We measured Krippendorff’s alpha \cite{Krippendorff2011ComputingKA} to determine inter-annotator agreement and found a moderate level of agreement (0.505\%) between the annotators. 

\noindent \textbf{Results} \,\, In this section, we present a detailed breakdown of the performance of the agentic framework against the direct prompting strategy across the different test sets. \Cref{tab:agentvsnonagentdet} presents the detailed results from the experiments. We also present our ablation study strategy in \Cref{tab:ablation_strat}.
\begin{table*}[t]
\centering
\caption{Automatic Evaluation results of generated stories (Agentic vs. Non-agentic) with pairwise additive prompting. Here, `I' in `LLaMA-3-Xb-I' stands for Instruction tuned versions, and `Agentic' and `Direct' stands for Agentic framework and Direct prompting strategy respectively. We calculate the \% of wins for these two different strategies and report them in this table. The \textcolor{gray}{\textbf{Gray}} text indices the number of samples for each case.}
\label{tab:agentvsnonagentdet}
\resizebox{\textwidth}{!}{%
\begin{tabular}{lcccc|cccc|cccc}
                               & \multicolumn{4}{c}{\textbf{Pew}}          & \multicolumn{4}{c}{\textbf{Tableau}}     & \multicolumn{4}{c}{\textbf{Gapminder}}  \\ \Xhline{1pt}
Model &
  Samples &
  \begin{tabular}[c]{@{}c@{}}Agentic\\ Win (\%)\end{tabular} &
  \begin{tabular}[c]{@{}c@{}}Direct\\ Win (\%)\end{tabular} &
  Tie (\%) &
  Samples &
  \begin{tabular}[c]{@{}c@{}}Agentic\\ Win (\%)\end{tabular} &
  \begin{tabular}[c]{@{}c@{}}Direct\\ Win (\%)\end{tabular} &
  Tie (\%) &
  Samples &
  \begin{tabular}[c]{@{}c@{}}Agentic\\ Win (\%)\end{tabular} &
  \begin{tabular}[c]{@{}c@{}}Direct\\ Win (\%)\end{tabular} &
  Tie (\%) \\ \cmidrule{1-13}
\multirow{2}{*}{GPT-4o}        & \multirow{2}{*}{321} & 78.50 & 19.63 & 1.87 & \multirow{2}{*}{13} & 69.23 & 30.77 & 0 & \multirow{2}{*}{5} & 80.00 & 20.00 & 0 \\
                               &                      & \textcolor{gray}{252} & \textcolor{gray}{63} & \textcolor{gray}{6} &                     & \textcolor{gray}{9} & \textcolor{gray}{4} & \textcolor{gray}{0} &                    & \textcolor{gray}{4} & \textcolor{gray}{1} & \textcolor{gray}{0} \\
\multirow{2}{*}{LLaMA-3-8b-I}  & \multirow{2}{*}{321} & 40.81 & 55.45 & 3.74 & \multirow{2}{*}{13} & 53.85 & 38.46 & 7.69 & \multirow{2}{*}{5} & 60 & 40 & 0 \\
                               &                      & \textcolor{gray}{131} & \textcolor{gray}{178} & \textcolor{gray}{12} &                     & \textcolor{gray}{7} & \textcolor{gray}{5} & \textcolor{gray}{1} &                    & \textcolor{gray}{3} & \textcolor{gray}{2} & \textcolor{gray}{0} \\
\multirow{2}{*}{LLaMA-3-70b-I} & \multirow{2}{*}{321} & 58.25 & 40.19 & 1.56 & \multirow{2}{*}{13} & 69.23 & 30.77 & 0 & \multirow{2}{*}{5} & 60 & 40 & 0 \\
                               &                      & \textcolor{gray}{187} & \textcolor{gray}{129} & \textcolor{gray}{5} &                     & \textcolor{gray}{9} & \textcolor{gray}{4} & \textcolor{gray}{0} &                    & \textcolor{gray}{3} & \textcolor{gray}{2} & \textcolor{gray}{0} \\ \Xhline{1pt}
\end{tabular}%
}
\end{table*}

\section{Additional Error Analysis}
\begin{figure*}[t]
     \centering
        % \vspace{mm}
        \includegraphics[width=\textwidth]{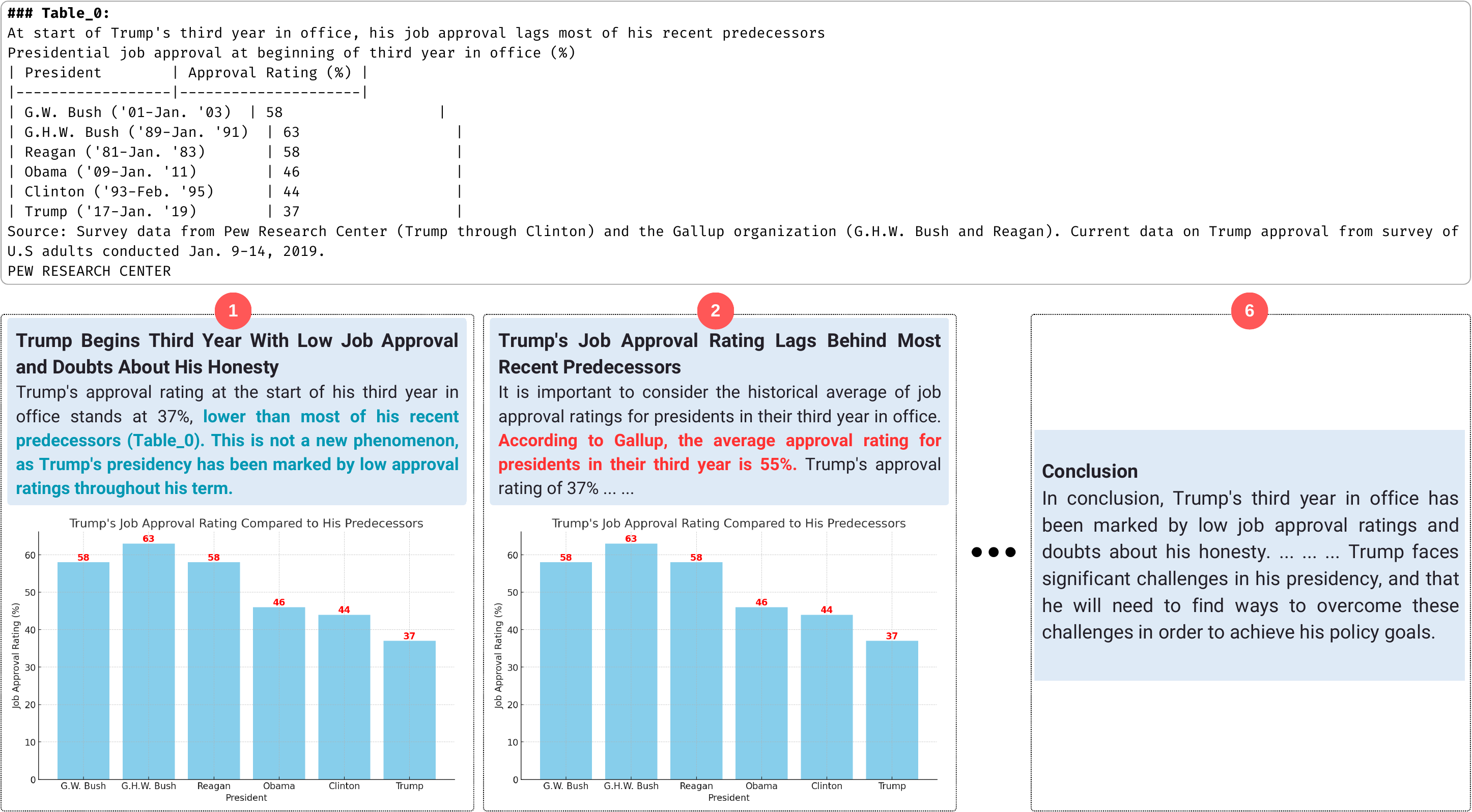}%emnlp2020-templates/Figures/Pipeline.png}
        % \vspace{-3mm}
         \caption{
         Examples of Factual and Hallucination errors in LLaMA-3-8b-instruct generated story using the Agentic framework. Here, \textcolor{moonstoneblue}{\textbf{Blue}} color denotes hallucinated text, and \textcolor{red}{\textbf{Red}} color denotes text containing factual errors.
         % \vspace{-3mm}
          }
    \label{fig:fact_hall_error}
\end{figure*}

\begin{figure*}[t]
     \centering
        % \vspace{mm}
        \includegraphics[width=\textwidth]{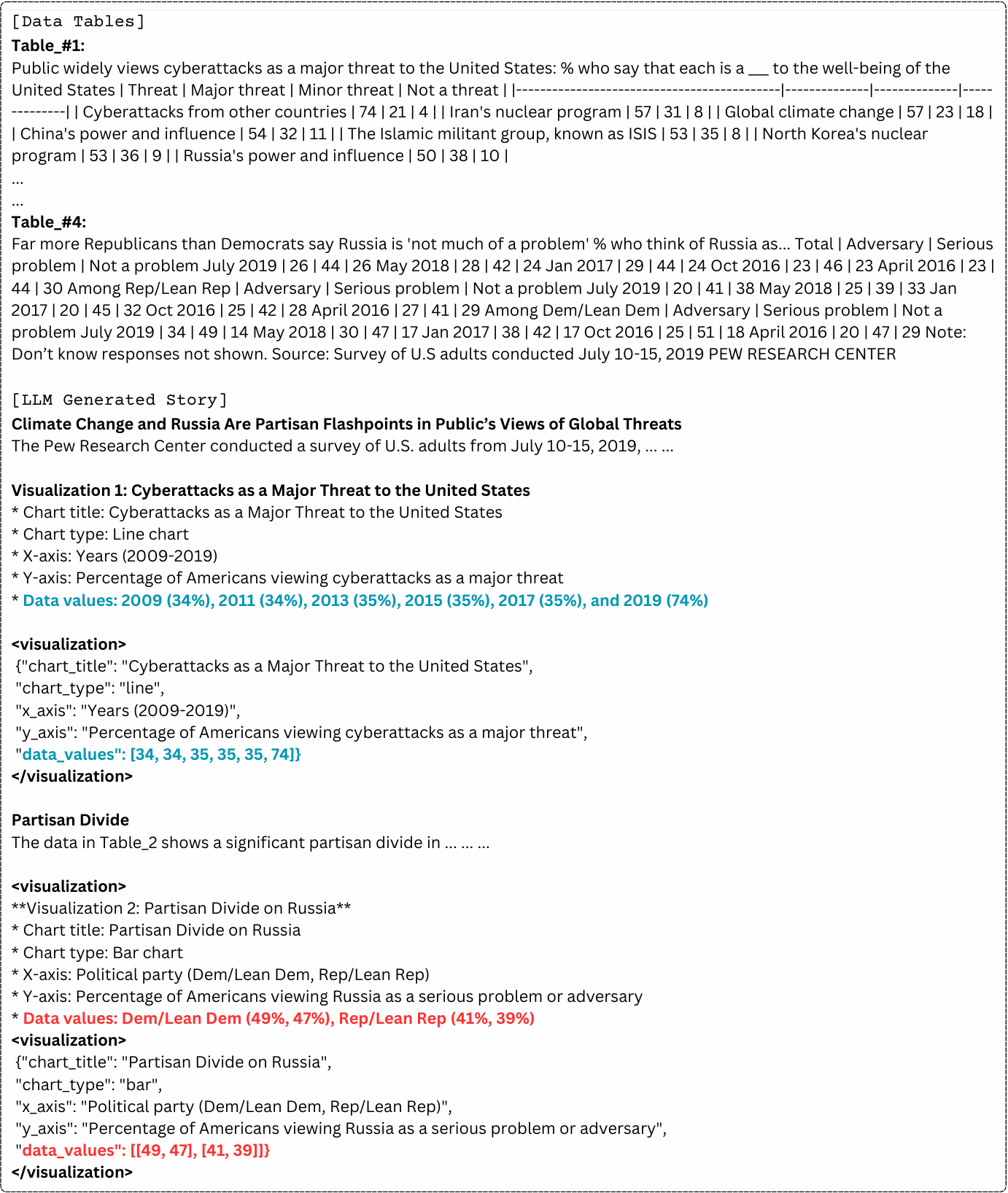}%emnlp2020-templates/Figures/Pipeline.png}
        % \vspace{-3mm}
         \caption{
         Examples of Factual and Hallucination errors in LLaMA-3-8b-instruct generated story using the Agentic framework. Here, \textcolor{moonstoneblue}{\textbf{Blue}} color denotes hallucinated text, and \textcolor{red}{\textbf{Red}} color denotes text containing factual errors.
         % \vspace{-3mm}
          }
    \label{fig:fact_hall_error_original}
\end{figure*}

In this section, we present examples of errors that occurred in the generated stories. For instance, \Cref{fig:fact_hall_error} illustrates a story generated by the LLaMA-3-8b-instruct model where factual errors are in `\textbf{Section 2}' where it mentions `\textit{average approval rating for presidents in the third year is 55\%}' according to the `\textbf{Table\_\#0}' in the figure, however, it is actually less than 55\% (the average is 53.8\%). Furthermore, we found that most factual error occurs in the `\textit{Visualization Specifications}' as exemplified by \Cref{fig:fact_hall_error_original}. Additionally, hallucinating data values is another concern at the time of narration generation, even though verification steps are included at each stage of the agentic framework. One such case is illustrated in \Cref{fig:fact_hall_error}, where the LLaMA-3-8b-instruct model hallucinated facts such as `\textit{Trump's presidency has been marked by low approval ratings throughout his term}', whereas the data in the table only gives a picture of first three years. Similar to the factual errors, most of the hallucinations are prevalent in the `\textit{Visualization Specifications}' like \Cref{fig:fact_hall_error_original}.

\begin{figure*}[t]
     \centering
        % \vspace{mm}
        \includegraphics[width=\textwidth]{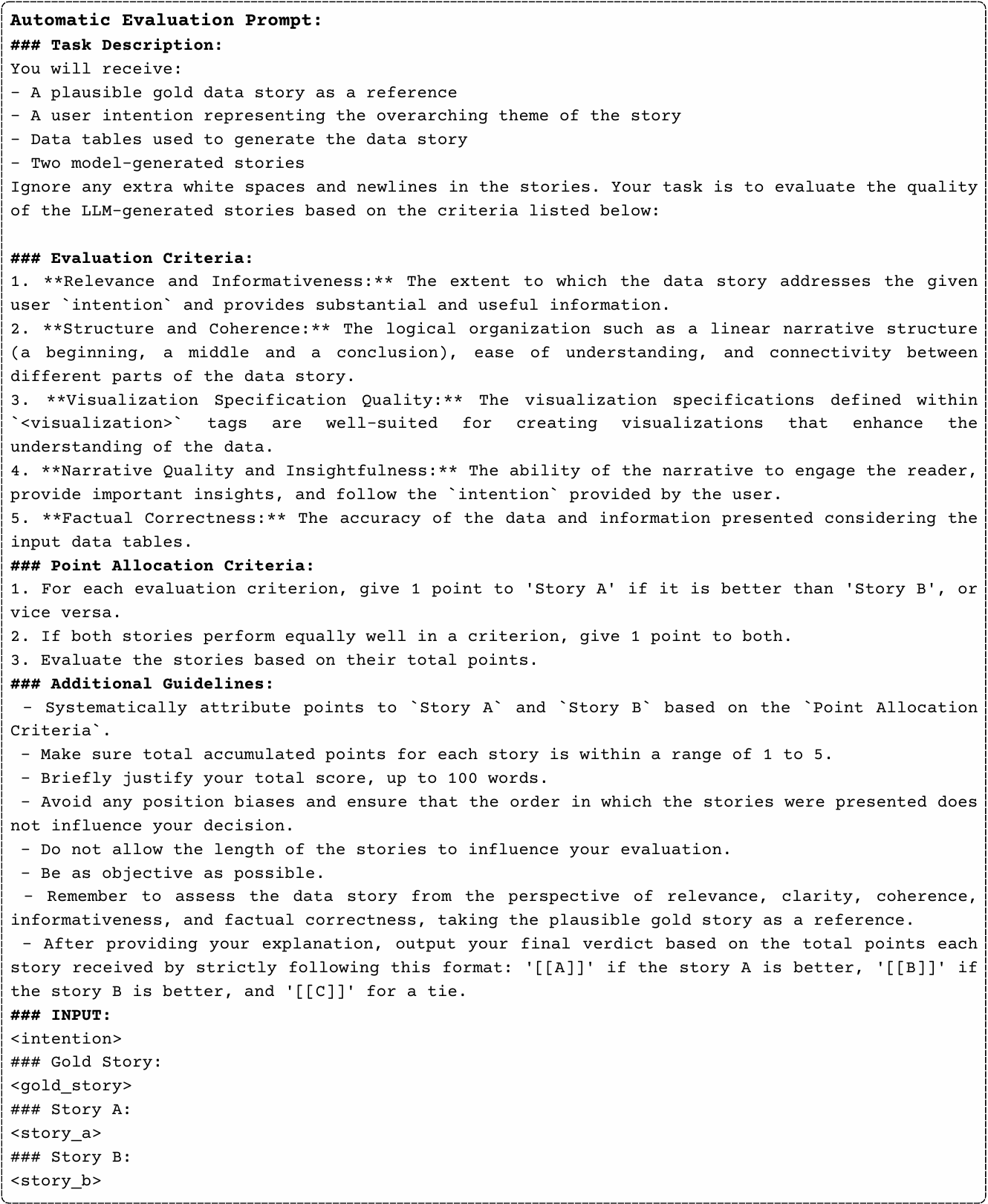}%emnlp2020-templates/Figures/Pipeline.png}
        % \vspace{-3mm}
         \caption{
         Pairwise automatic evaluation prompt.
         % \vspace{-3mm}
          }
    \label{fig:auto_eval_instruction}
\end{figure*}

\begin{figure*}[t]
     \centering
        % \vspace{mm}
        \includegraphics[width=\textwidth]{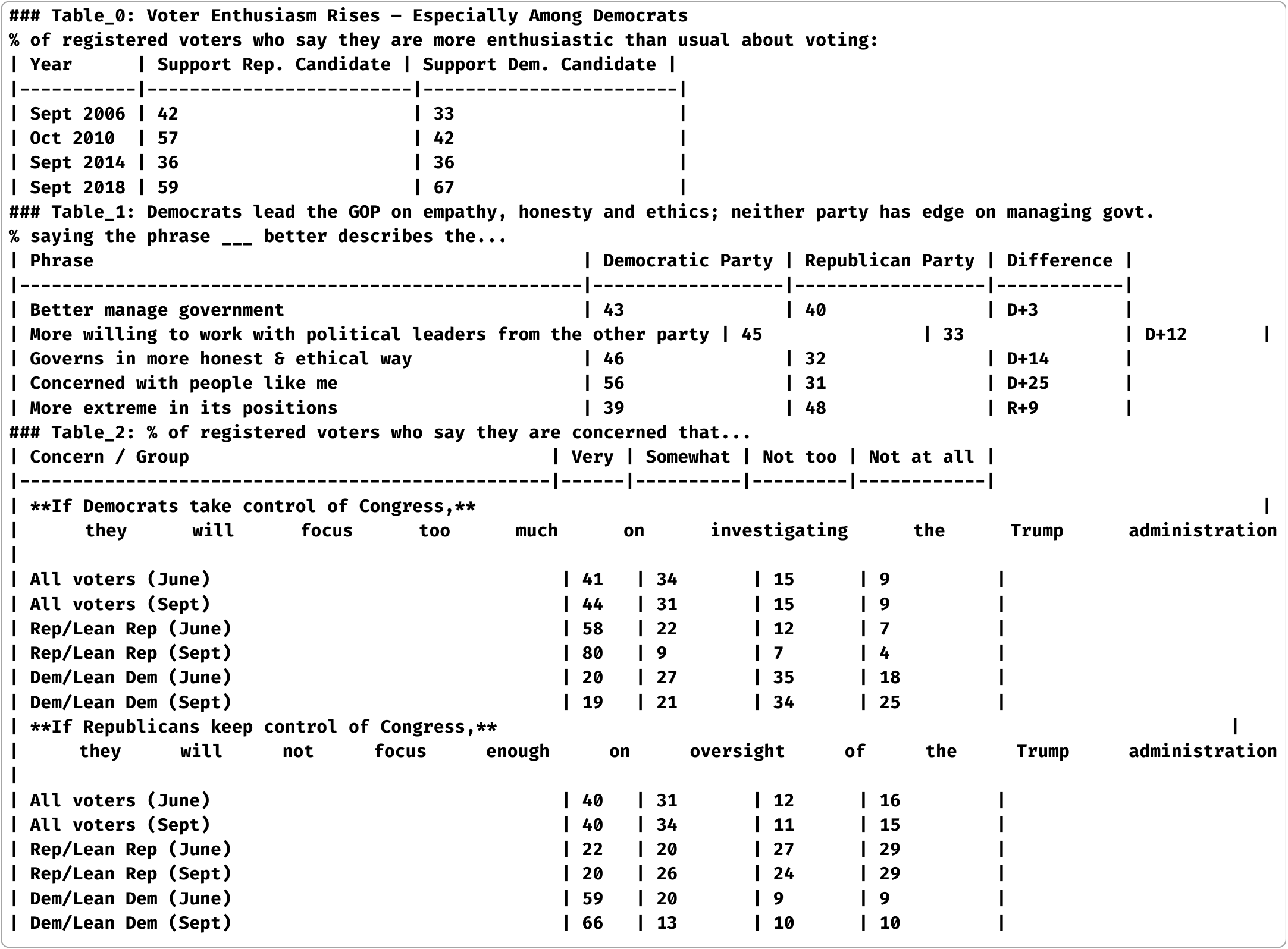}%emnlp2020-templates/Figures/Pipeline.png}
        % \vspace{-3mm}
         \caption{
         The tables corresponding to \Cref{fig:fact_hall_gpt4o}.
         % \vspace{-3mm}
          }
    \label{fig:fact_hall_gpt4o_table}
\end{figure*}

\begin{figure*}[t]
     \centering
        % \vspace{mm}
        \includegraphics[width=\textwidth]{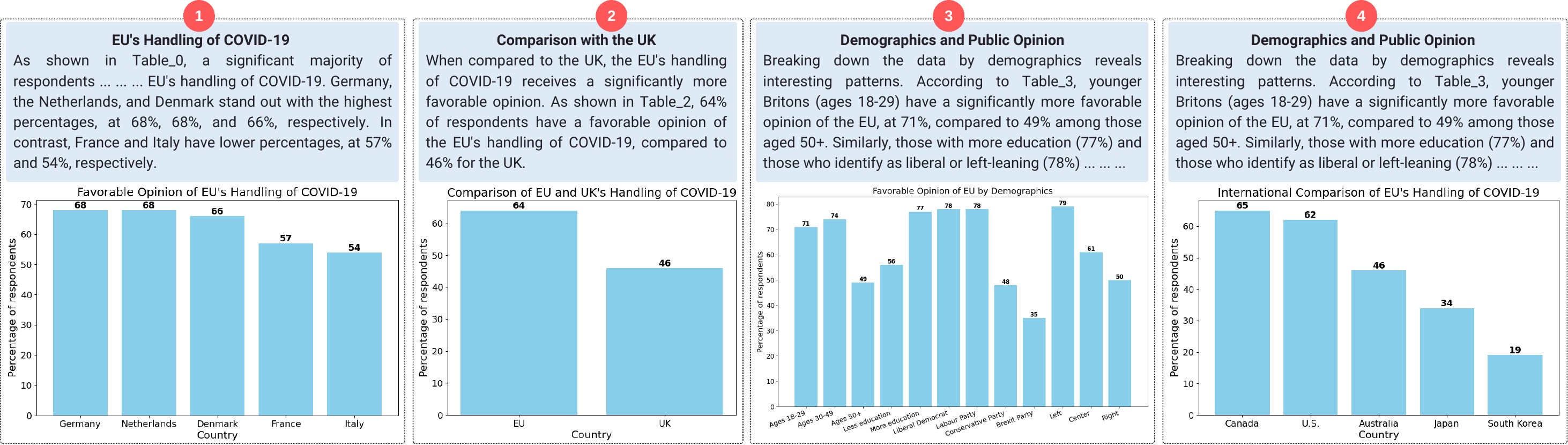}%emnlp2020-templates/Figures/Pipeline.png}
        % \vspace{-3mm}
         \caption{
         A figure demonstrating the `Coherence' issue of the LLaMA-3-8b model.
         % \vspace{-3mm}
          }
    \label{fig:coher_8b}
\end{figure*}

\section{Examples}
%%% GOLD STORY EXAMPLE (PEW)
\begin{figure*}[t]
     \centering
        % \vspace{mm}
        \includegraphics[width=\textwidth]{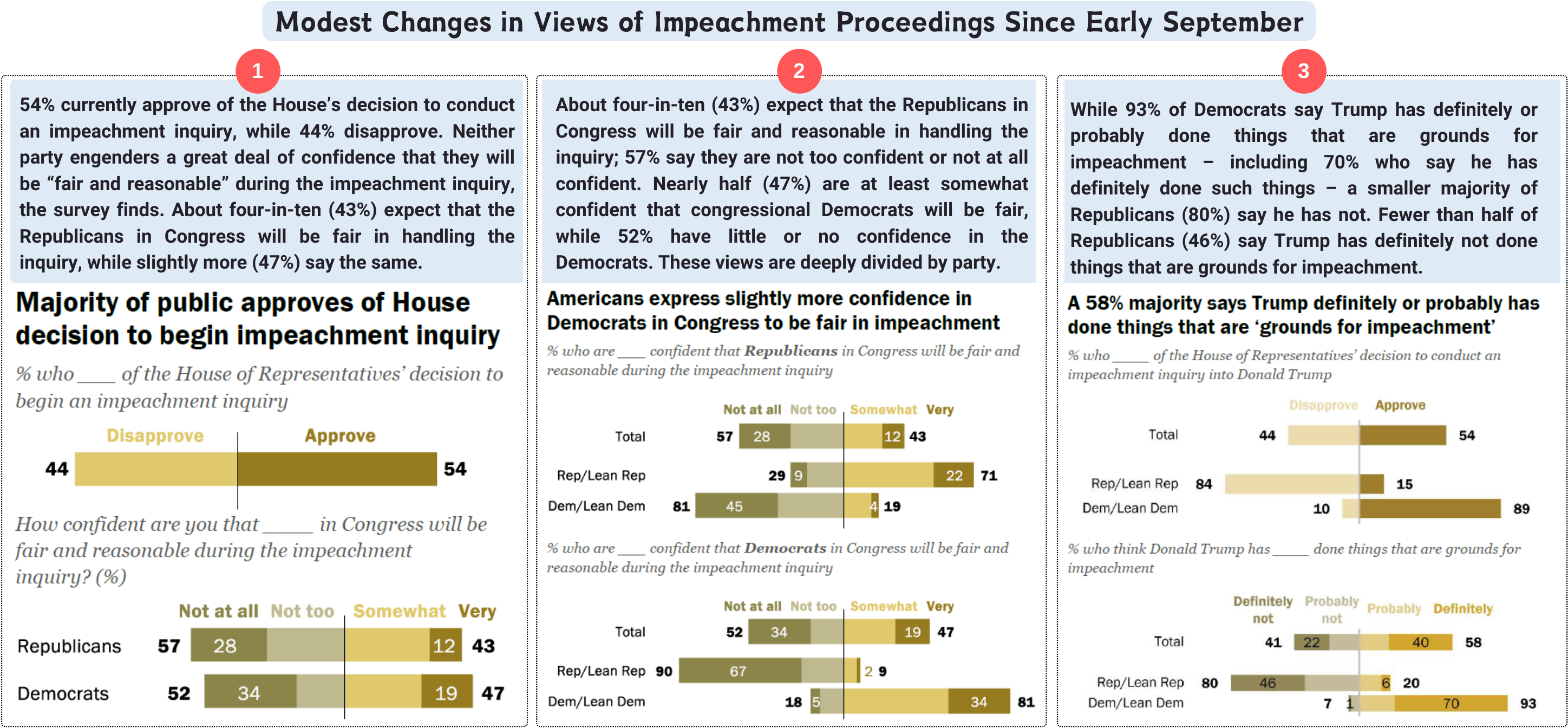}%emnlp2020-templates/Figures/Pipeline.png}
        % \vspace{-3mm}
         \caption{
         An example data story in our corpus collected
from Pew \cite{pewresearch}. 
         % \vspace{-3mm}
          }
    \label{fig:pew_data_story}
\end{figure*} 

%%% GOLD STORY EXAMPLE (TABLEAU)
\begin{figure*}[t]
     \centering
        % \vspace{mm}
        \includegraphics[width=\textwidth]{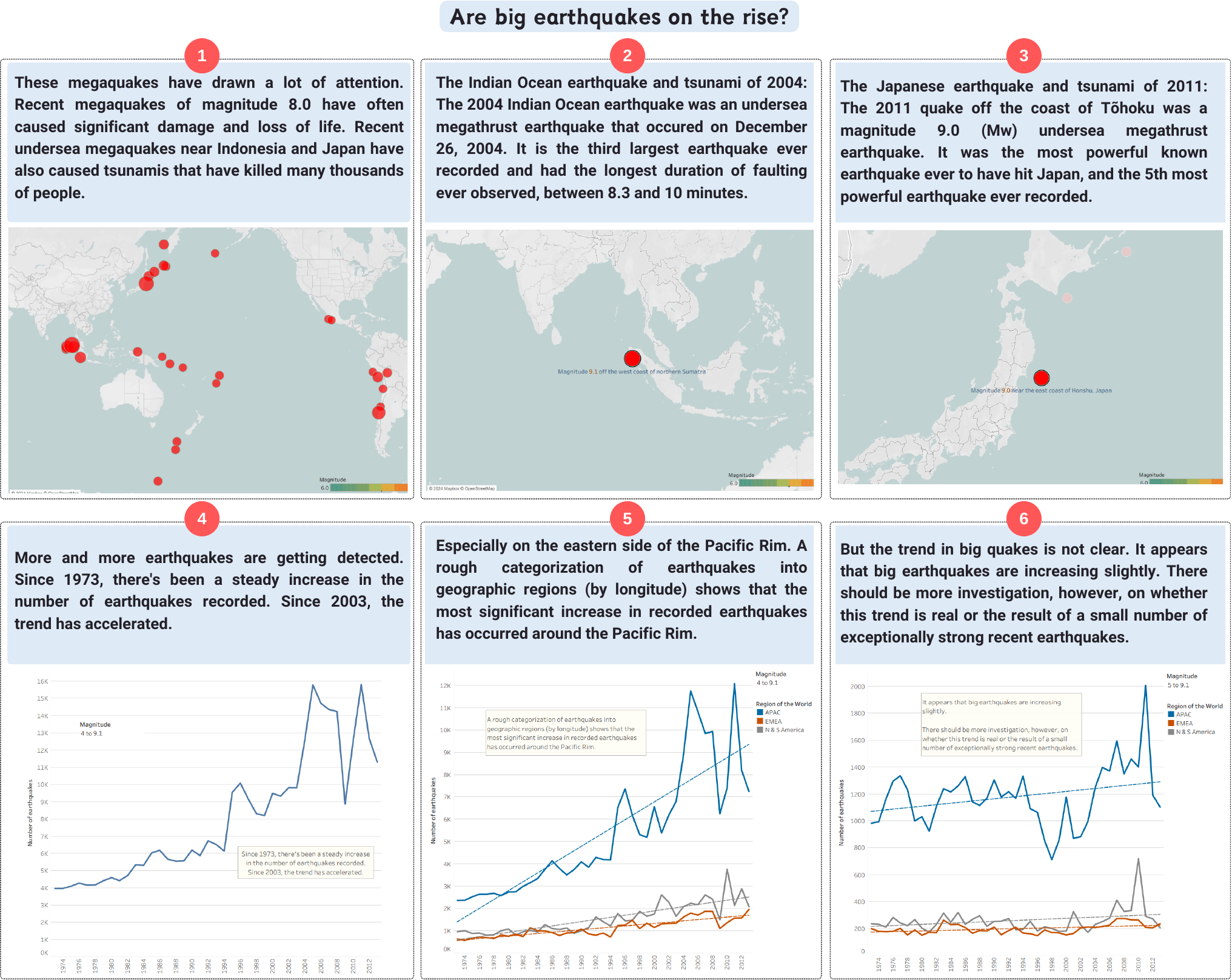}%emnlp2020-templates/Figures/Pipeline.png}
        % \vspace{-3mm}
         \caption{
         An example data story in our corpus collected
from Tableau \cite{tableaupublic}.
         % \vspace{-3mm}
          }
    \label{fig:tableau_data_story}
\end{figure*} 

%%% Initial Reflection Prompt
\begin{figure*}[t!]
     \centering
        % \vspace{mm}
        \includegraphics[width=\textwidth]{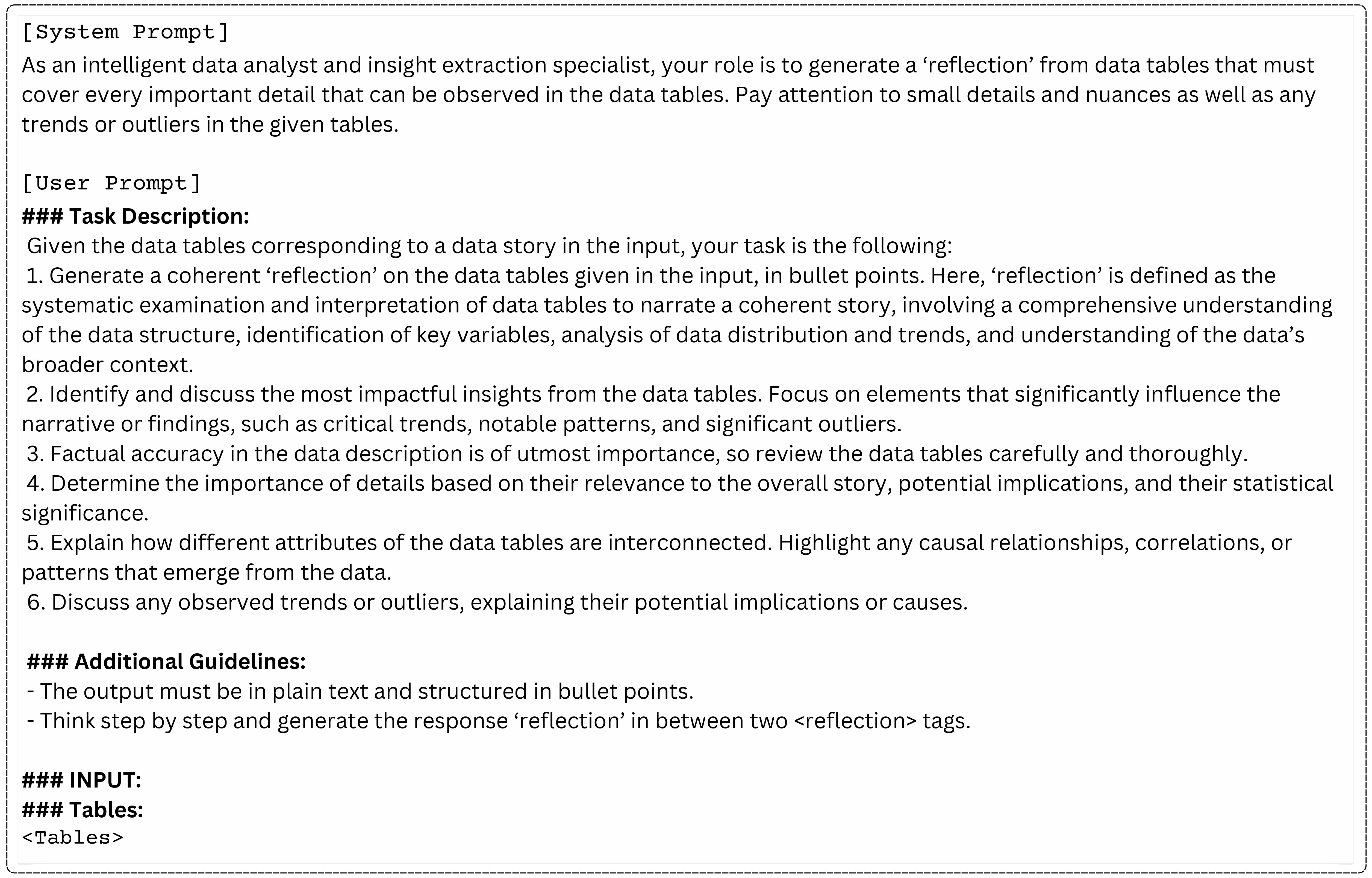}%emnlp2020-templates/Figures/Pipeline.png}
        % \vspace{-3mm}
         \caption{
         The figure presents the prompt used to generate the initial `Reflection'.
         % \vspace{-3mm}
          }
    \label{fig:init_refl}
\end{figure*} 

%%% Reflection Revision Prompt
\begin{figure*}[t!]
     \centering
        % \vspace{mm}
        \includegraphics[width=\textwidth]{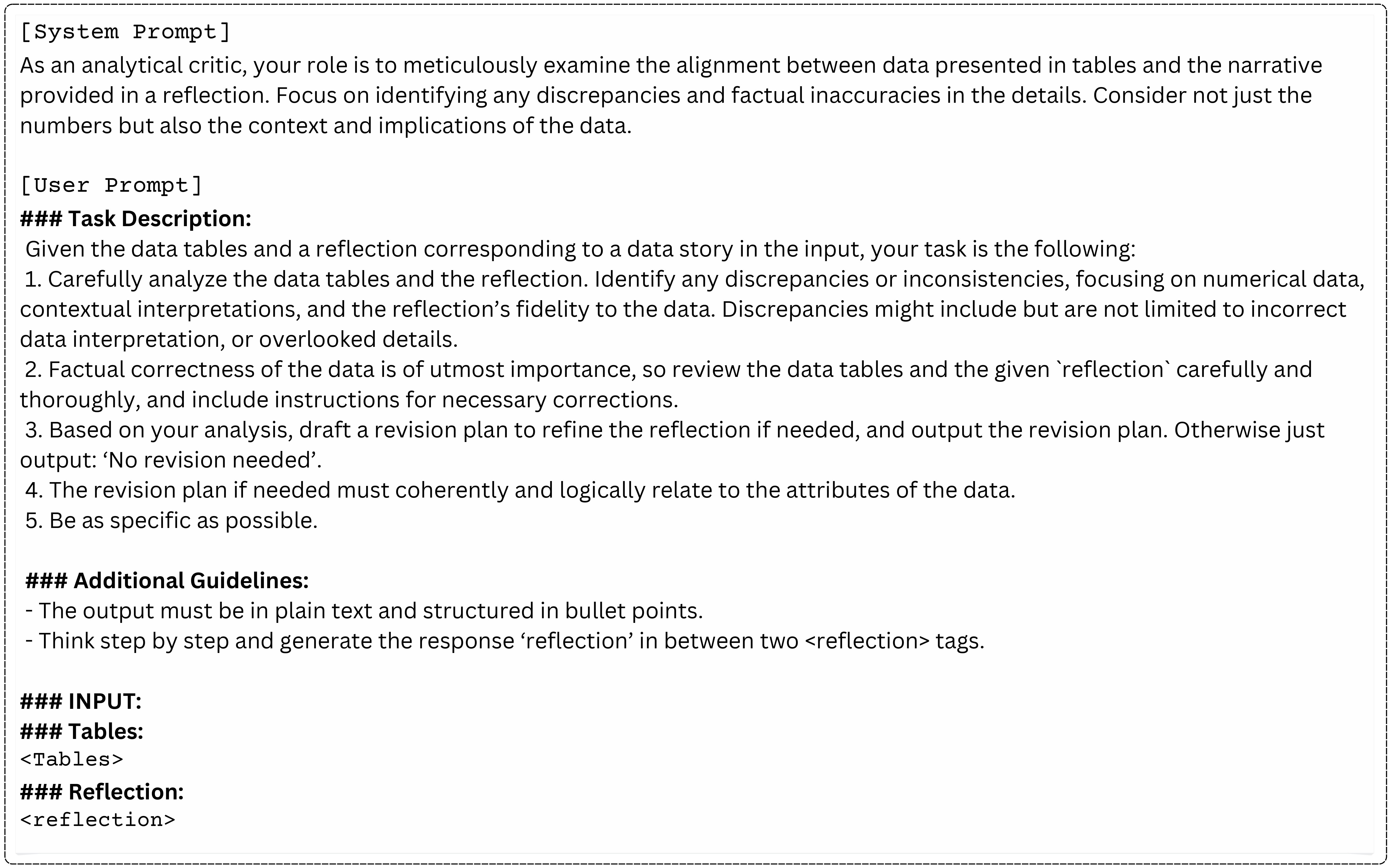}%emnlp2020-templates/Figures/Pipeline.png}
        % \vspace{-3mm}
         \caption{
         The figure presents the prompt used to generate the `Reflection' revision plan. 
         % \vspace{-3mm}
          }
    \label{fig:refl_rev_plan}
\end{figure*}

%%% Revised Reflection Prompt
\begin{figure*}[t!]
     \centering
        % \vspace{mm}
        \includegraphics[width=\textwidth]{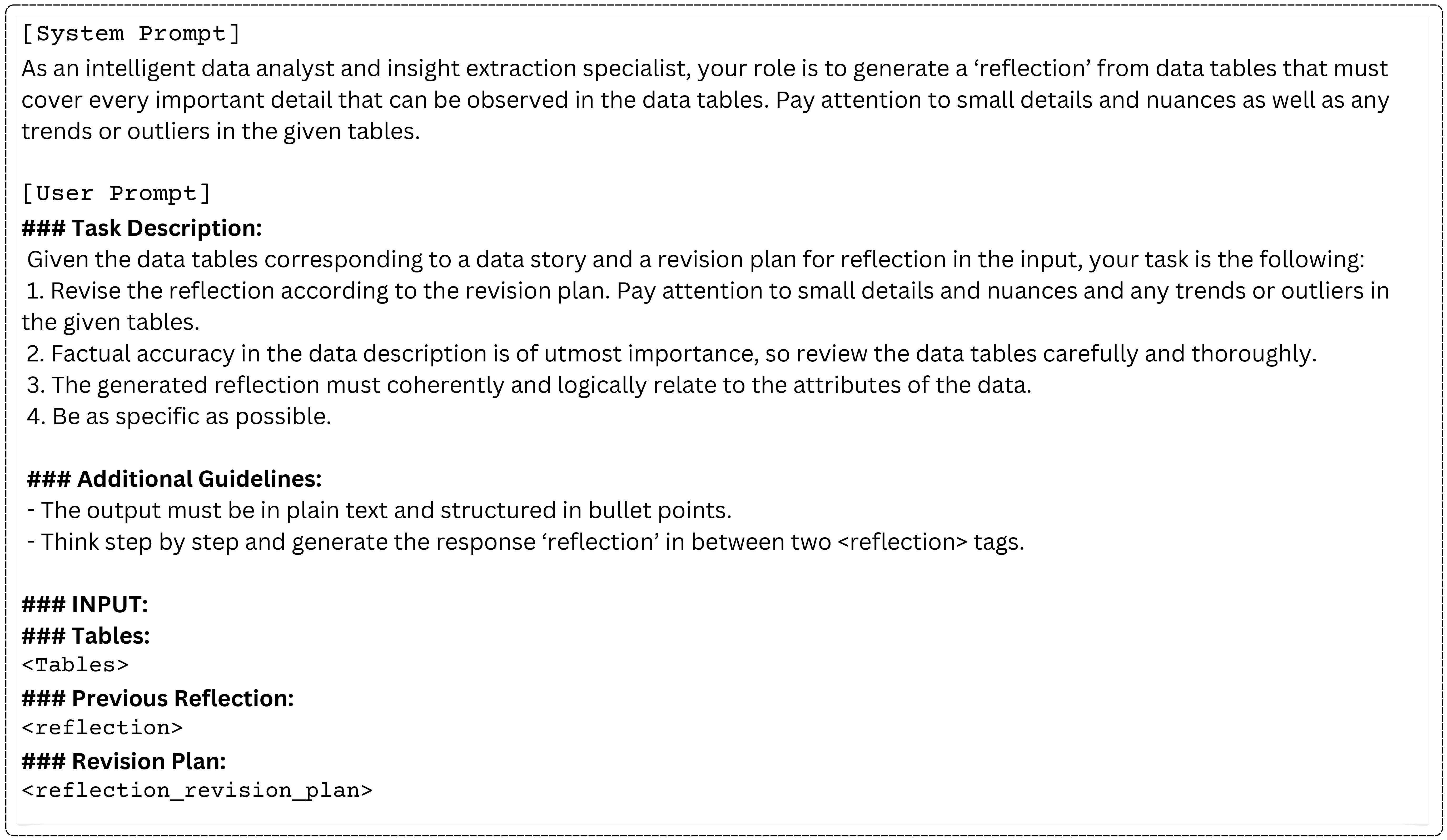}%emnlp2020-templates/Figures/Pipeline.png}
        % \vspace{-3mm}
         \caption{
         The figure presents the prompt used to generate the revised `Reflection'. 
         % \vspace{-3mm}
          }
    \label{fig:rev_refl}
\end{figure*}

%%% Initial Outline Prompt
\begin{figure*}[t!]
     \centering
        % \vspace{mm}
        \includegraphics[width=\textwidth]{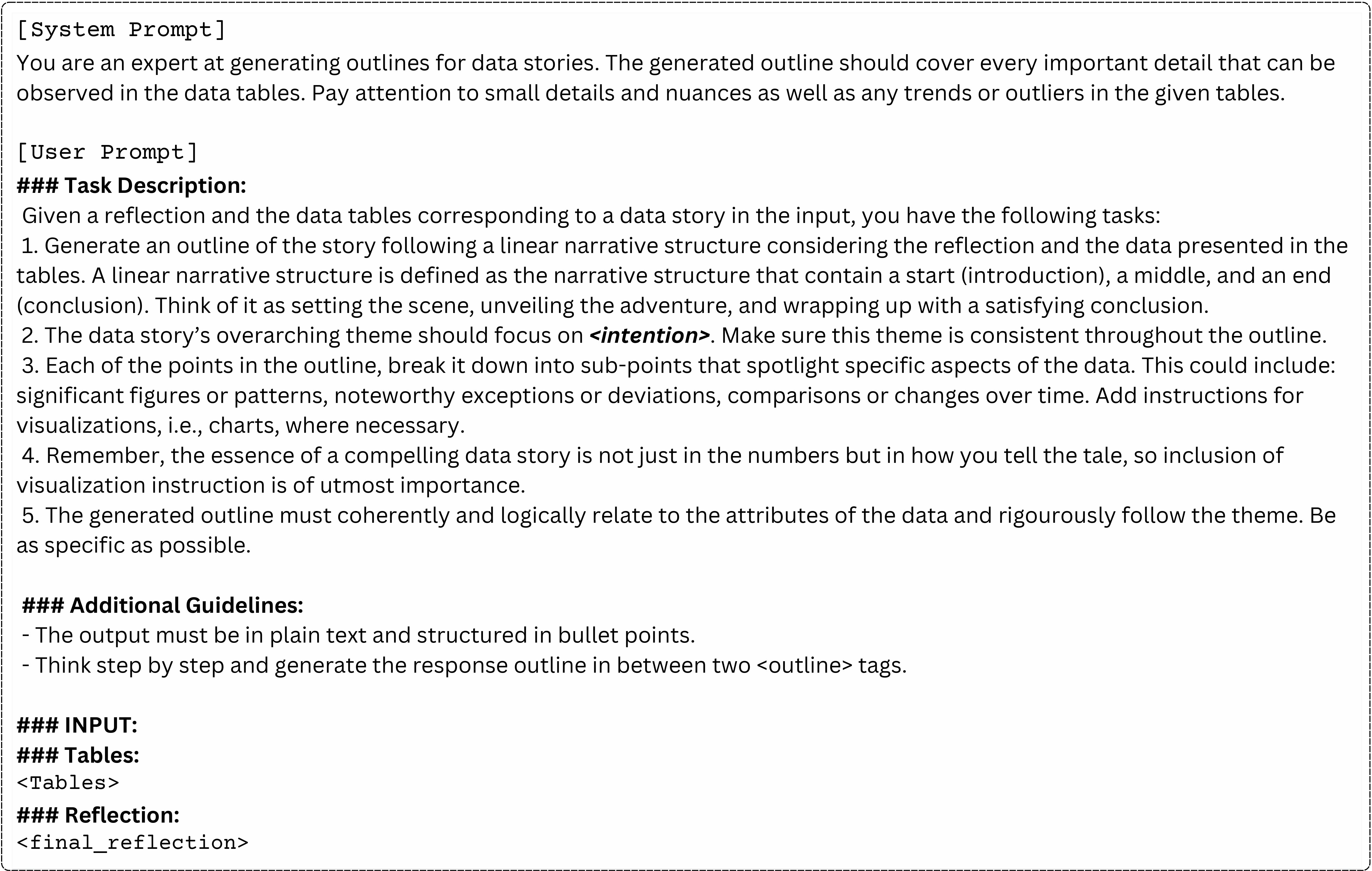}%emnlp2020-templates/Figures/Pipeline.png}
        % \vspace{-3mm}
         \caption{
         The figure presents the prompt used to generate the initial `Outline'.
         % \vspace{-3mm}
          }
    \label{fig:init_outl}
\end{figure*} 

%%% Outline Revision Prompt
\begin{figure*}[t!]
     \centering
        % \vspace{mm}
        \includegraphics[width=\textwidth]{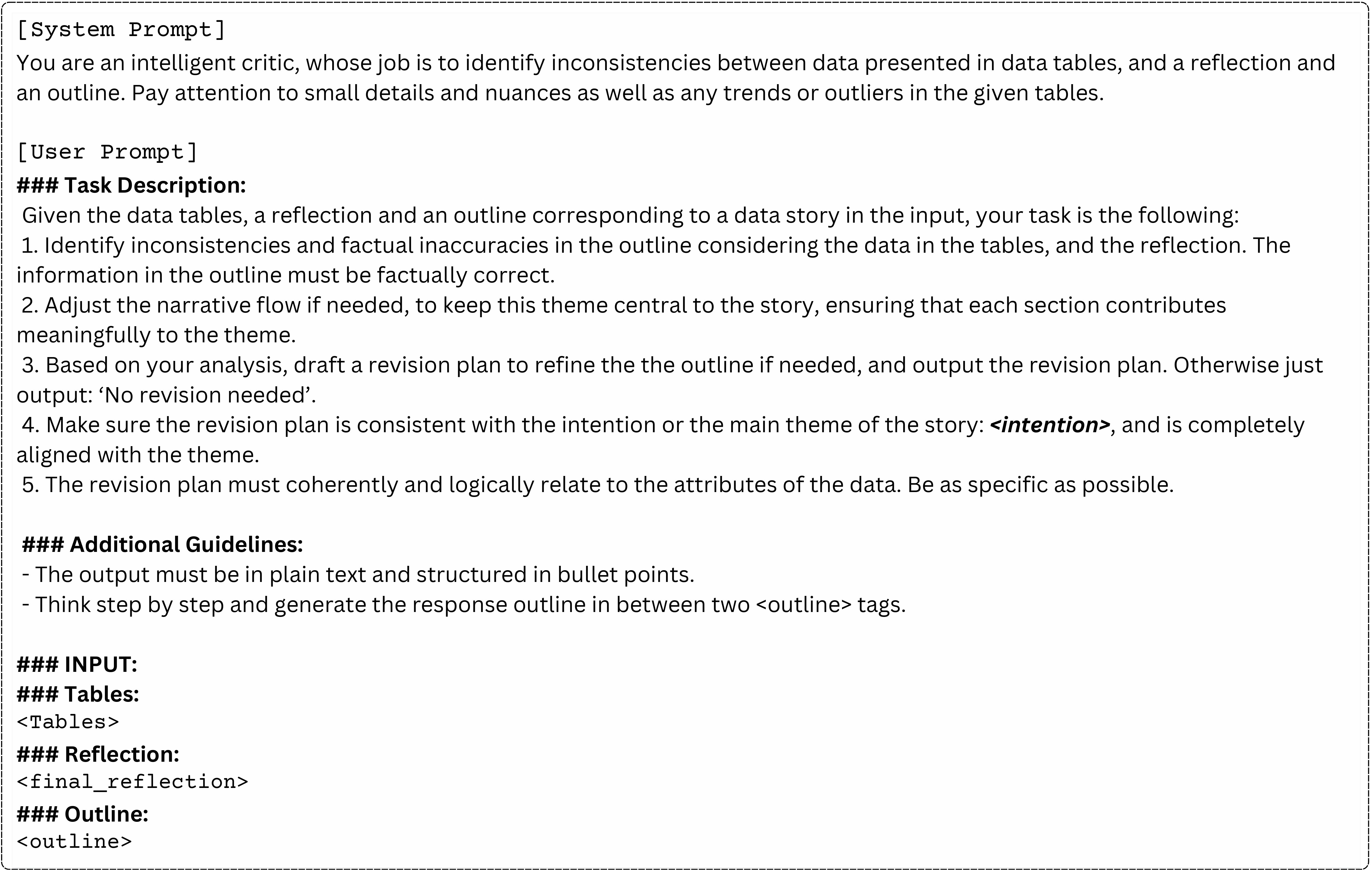}%emnlp2020-templates/Figures/Pipeline.png}
        % \vspace{-3mm}
         \caption{
         The figure presents the prompt used to generate the `Outline' revision plan. 
         % \vspace{-3mm}
          }
    \label{fig:outl_rev_plan}
\end{figure*}

%%% Revised Outline Prompt
\begin{figure*}[t!]
     \centering
        % \vspace{mm}
        \includegraphics[width=\textwidth]{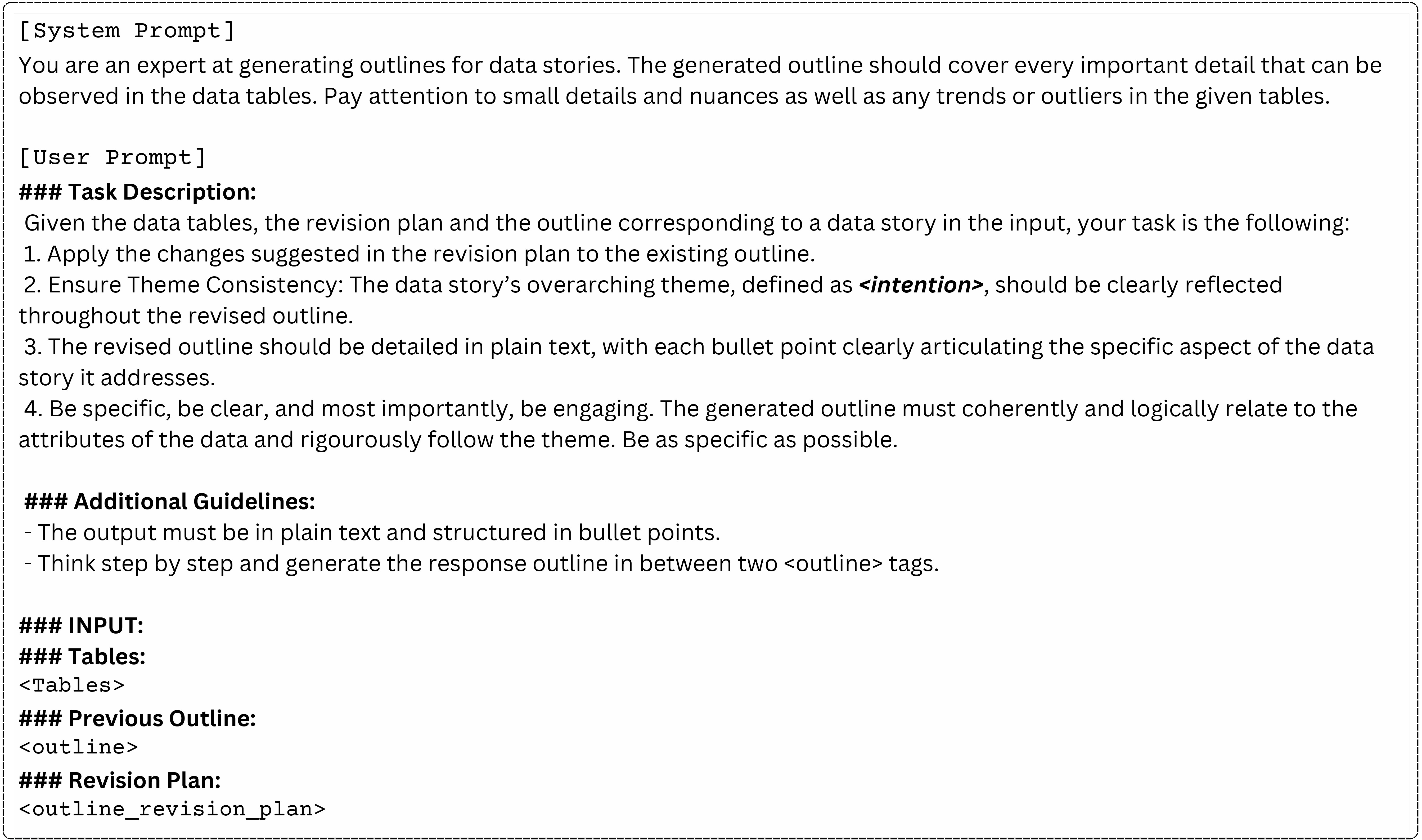}%emnlp2020-templates/Figures/Pipeline.png}
        % \vspace{-3mm}
         \caption{
         The figure presents the prompt used to generate the revised `Outline'. 
         % \vspace{-3mm}
          }
    \label{fig:rev_outl}
\end{figure*}

%%% Initial Narration Prompt
\begin{figure*}[t!]
     \centering
        % \vspace{mm}
        \includegraphics[width=\textwidth]{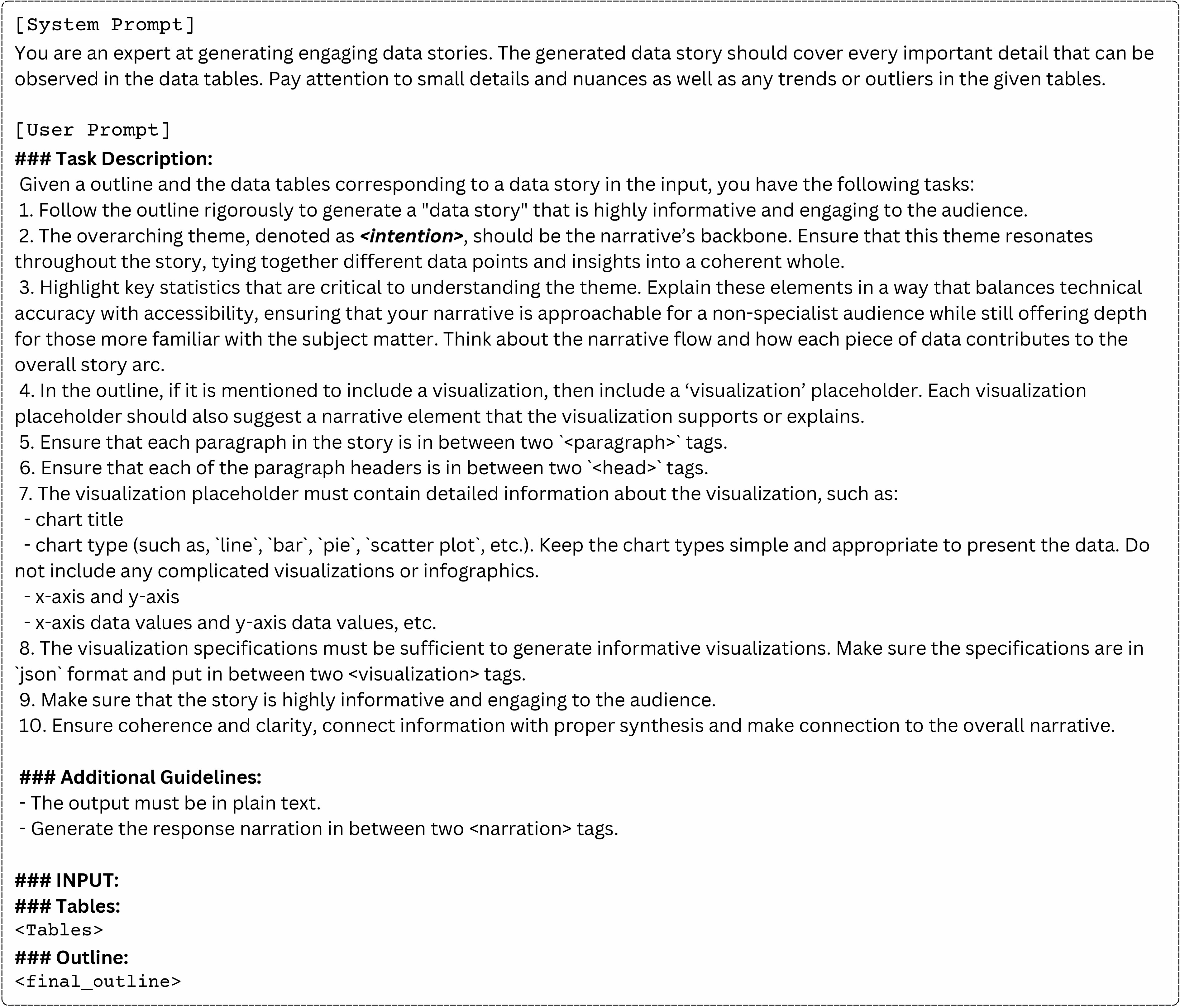}%emnlp2020-templates/Figures/Pipeline.png}
        % \vspace{-3mm}
         \caption{
         The figure presents the prompt used to generate the initial `Narration'.
         % \vspace{-3mm}
          }
    \label{fig:init_narr}
\end{figure*} 

%%% Narration Revision Prompt
\begin{figure*}[t!]
     \centering
        % \vspace{mm}
        \includegraphics[width=\textwidth]{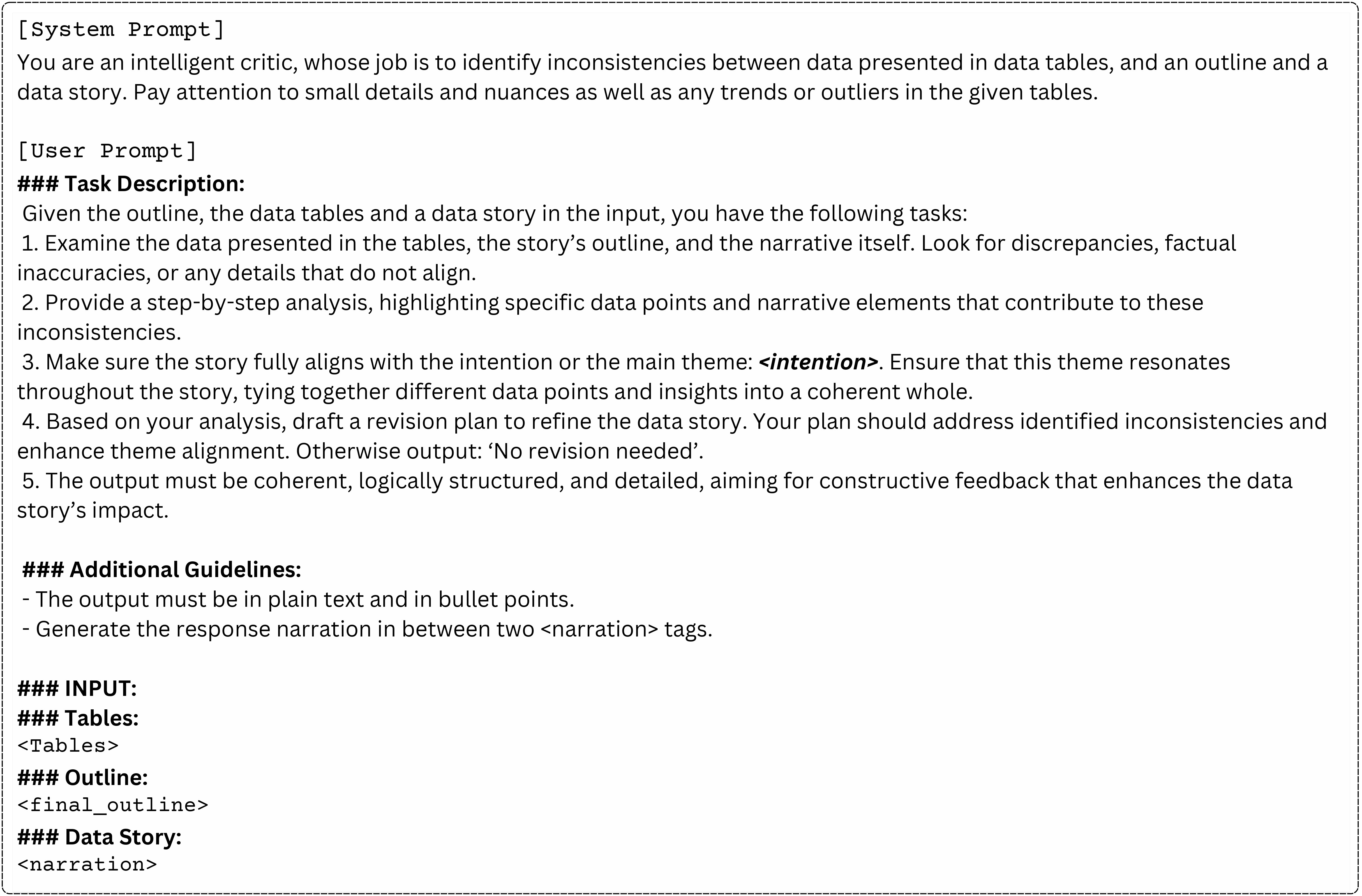}%emnlp2020-templates/Figures/Pipeline.png}
        % \vspace{-3mm}
         \caption{
         The figure presents the prompt used to generate the `Narration' revision plan. 
         % \vspace{-3mm}
          }
    \label{fig:narr_rev_plan}
\end{figure*}

%%% Revised Narration Prompt
\begin{figure*}[t!]
     \centering
        % \vspace{mm}
        \includegraphics[width=\textwidth]{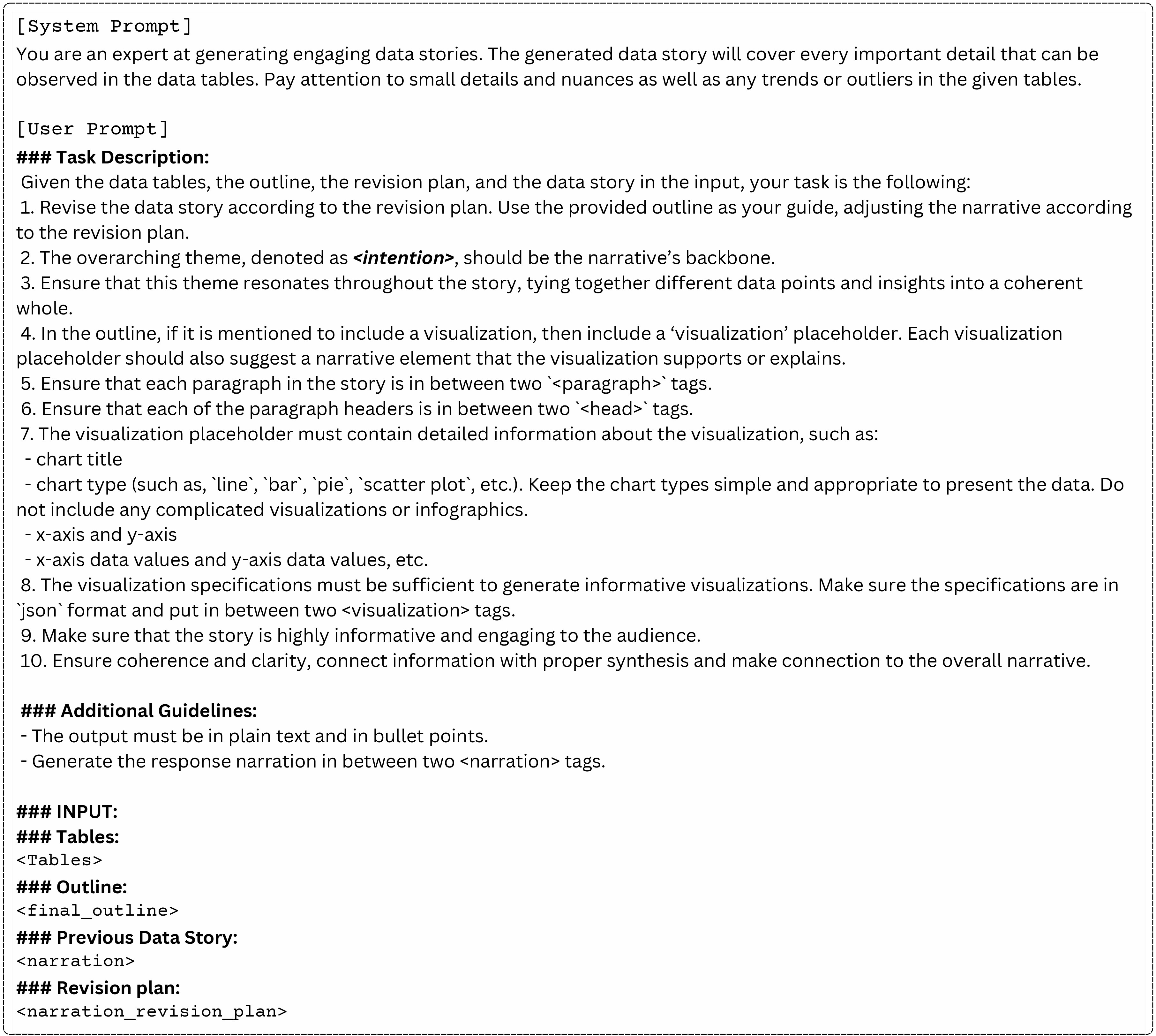}%emnlp2020-templates/Figures/Pipeline.png}
        % \vspace{-3mm}
         \caption{
         The figure presents the prompt used to generate the revised `Narration'. 
         % \vspace{-3mm}
          }
    \label{fig:rev_narr}
\end{figure*}

% %%% GOLD STORY EXAMPLE (Gapminder)
% \begin{figure*}[t!]
%      \centering
%         % \vspace{mm}
%         \includegraphics[width=\textwidth]{emnlp2020-templates/figures/example_data_story_gap.jpg}%emnlp2020-templates/Figures/Pipeline.png}
%         \vspace{-3mm}
%          \caption{
%          The figure demonstrates an example data story from our test set (Gapminder).
%          \vspace{-3mm}
%           }
%     \label{fig:gap_data_story}
% \end{figure*} 

%%%% EXAMPLE GPT-4o story with Vis
\begin{figure*}[t!]
     \centering
        % \vspace{mm}
        \includegraphics[width=\textwidth]{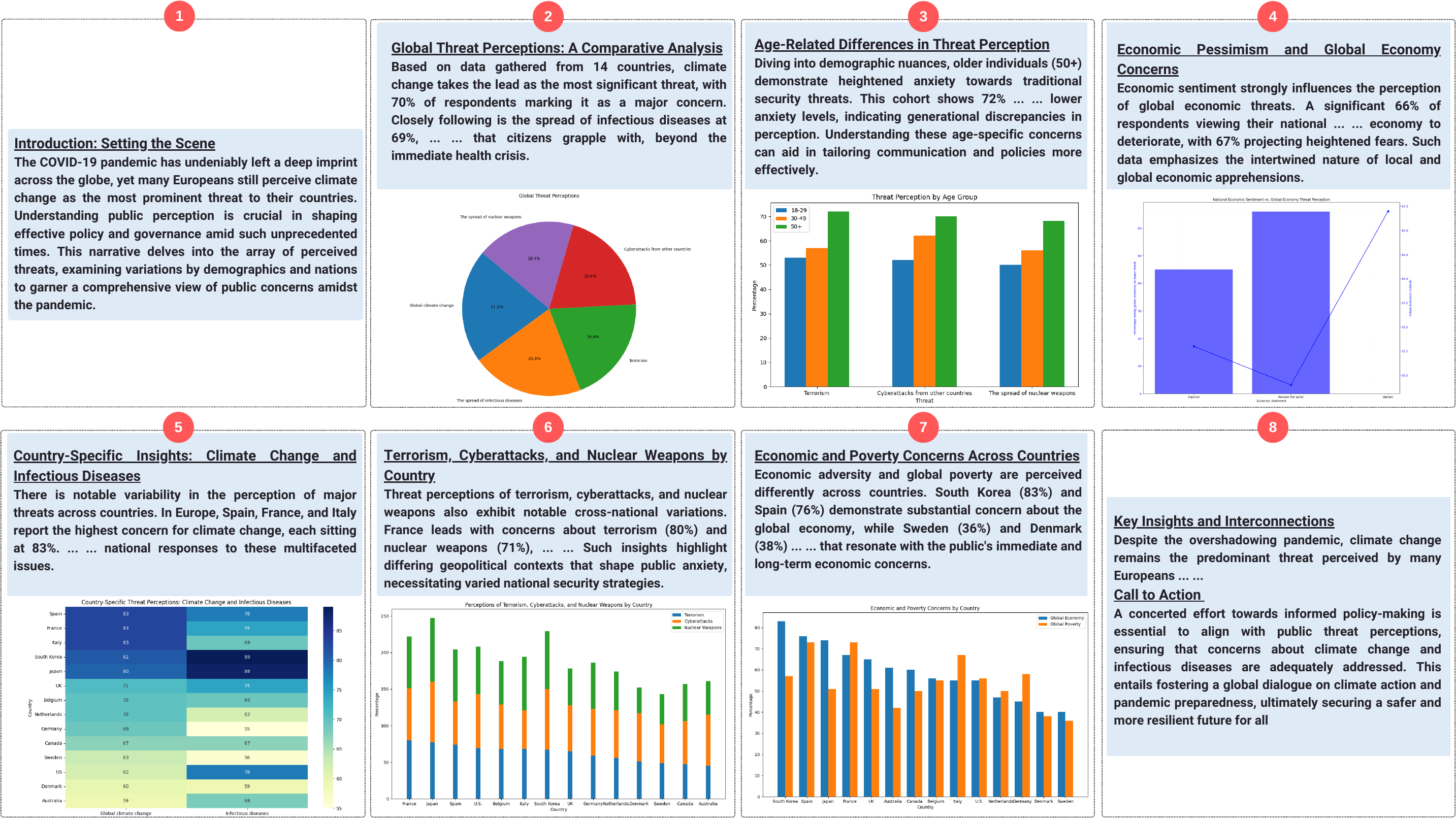}%emnlp2020-templates/Figures/Pipeline.png}
        % \vspace{-3mm}
         \caption{
         The figure demonstrates an example data story generated by GPT-4o using the agentic framework. Here, ‘...’ indicates abbreviated text for brevity. 
         % \vspace{-3mm}
          }
    \label{fig:gpt_story}
\end{figure*}

%%% EXAMPLE (GPT)
\begin{figure*}[t!]
     \centering
        % \vspace{mm}
        \includegraphics[width=\textwidth]{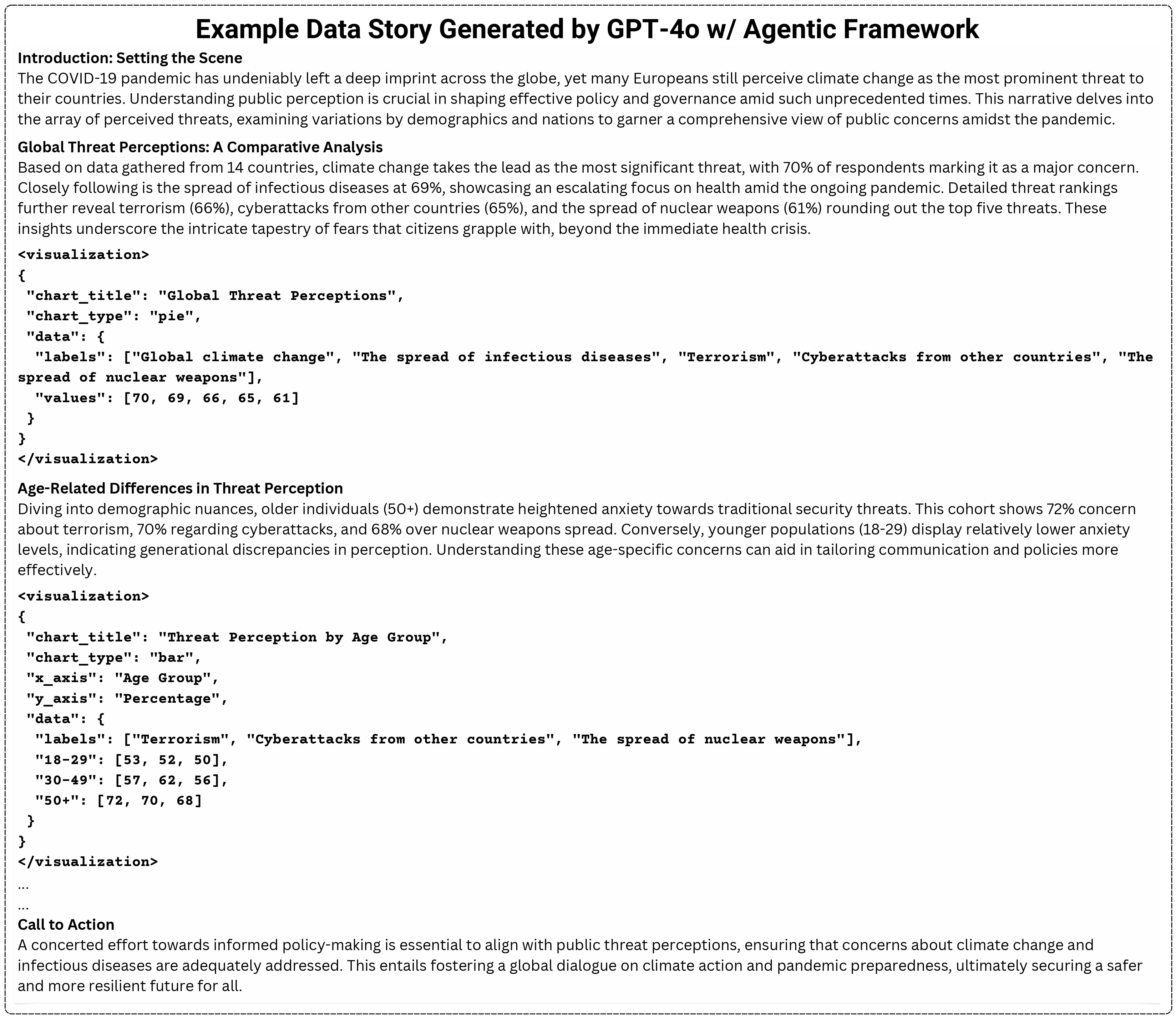}%emnlp2020-templates/Figures/Pipeline.png}
        % \vspace{-3mm}
         \caption{
         The figure demonstrates an example data story generated by GPT-4o in natural language text. Here, ‘...’ indicates abbreviated text for brevity.
         % \vspace{-3mm}
          }
    \label{fig:gpt_data_story_without_vis}
\end{figure*} 

%%% EXAMPLE (LLAMA-3)
\begin{figure*}[t!]
     \centering
        % \vspace{mm}
        \includegraphics[width=\textwidth]{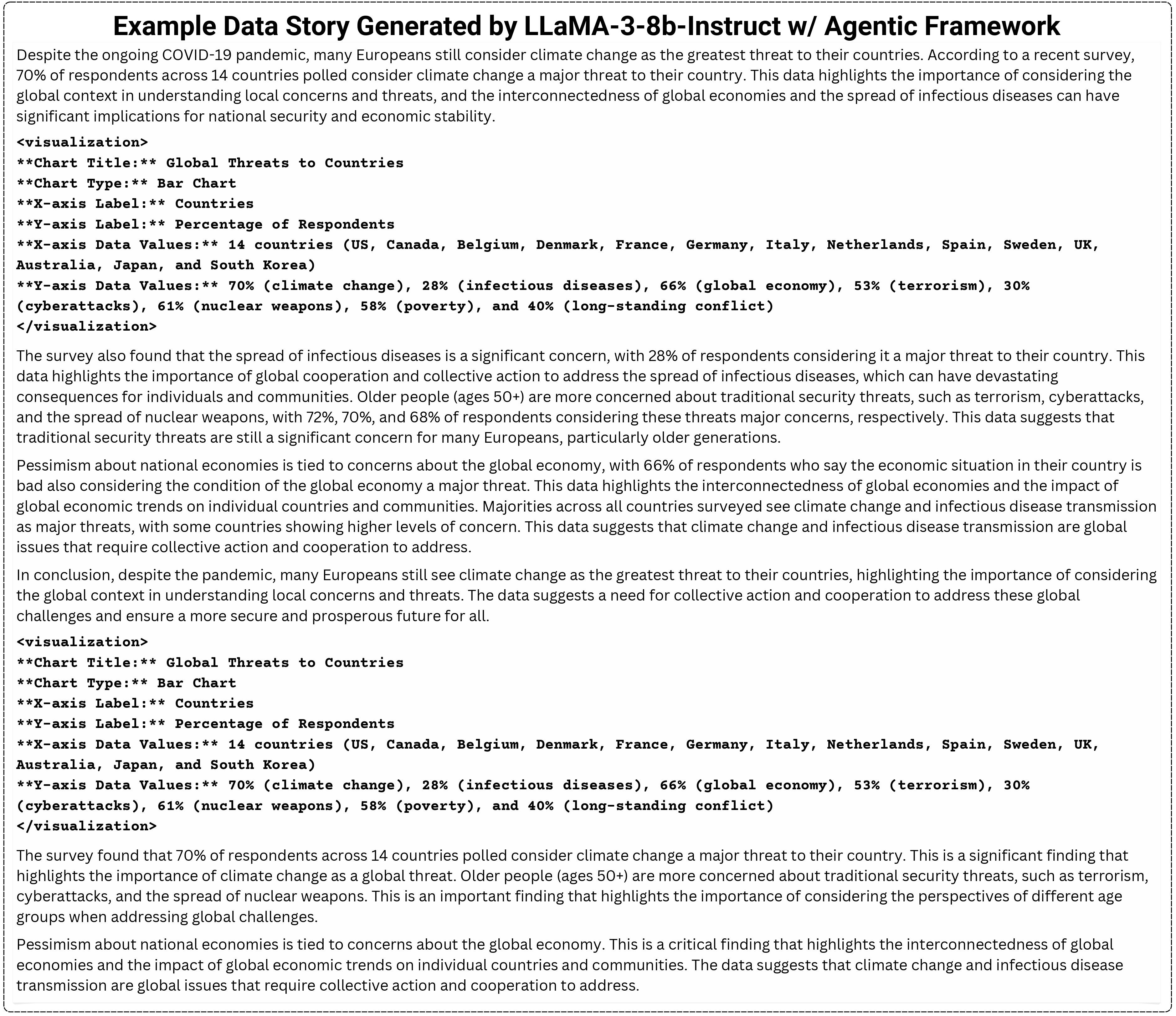}%emnlp2020-templates/Figures/Pipeline.png}
        % \vspace{-3mm}
         \caption{
         The figure demonstrates an example data story generated by the LLaMA-3-8b-instruct model in natural language text.   
         % \vspace{-3mm}
          }
    \label{fig:llama3_data_story_original}
\end{figure*} 

\end{appendices}

\end{document}